\documentclass[twocolumn]{svjour3}          %
\smartqed  %

\usepackage[dvipsnames, svgnames, x11names]{xcolor}
\usepackage[caption=false]{subfig}
\usepackage{graphicx}

\usepackage{float} 
\usepackage{multirow}
\usepackage{amssymb}
\usepackage{amsmath}
\usepackage[noend]{algpseudocode}
\usepackage{algorithmicx,algorithm}
\usepackage{natbib}
\usepackage{color}
\usepackage{pifont}
\usepackage{lmodern}
\usepackage{listings}
\usepackage{xspace}
\usepackage{adjustbox}
\usepackage{paralist} %
\usepackage{booktabs} %
\usepackage{enumitem}
\setitemize{noitemsep,topsep=0pt,parsep=0pt,partopsep=0pt}

\def\green#1{\textcolor{Green}{#1}}
\definecolor{orange}{rgb}{1,0.5,0}
\definecolor{citecolor}{RGB}{34,139,34}

\def\ie{i.e.,~}
\def\eg{e.g.,~}

\newcommand{\ra}[1]{\renewcommand{\arraystretch}{#1}}
\makeatletter\renewcommand\paragraph{\@startsection{paragraph}{4}{\z@}
	{.5em \@plus1ex \@minus.2ex}{-.5em}{\normalfont\normalsize\bfseries}}\makeatother
	
\newcommand{\tablestyle}[2]{\setlength{\tabcolsep}{#1}\renewcommand{\arraystretch}{#2}\centering\footnotesize}

\newlength\savewidth\newcommand\shline{\noalign{\global\savewidth\arrayrulewidth
		\global\arrayrulewidth 1pt}\hline\noalign{\global\arrayrulewidth\savewidth}}
\newcommand{\cmark}{\ding{51}}%

\usepackage[pagebackref=true,breaklinks=true,letterpaper=true,colorlinks,citecolor=citecolor,bookmarks=false]{hyperref}

\begin{document}

\title{BiSeNet V2: Bilateral Network with Guided Aggregation for Real-time Semantic Segmentation}

\author{Changqian Yu$^{1,2}$ \and
        Changxin Gao$^{1*}$ \and
        Jingbo Wang$^{3}$ \and
        Gang Yu$^{4}$ \and
        Chunhua Shen$^{2}$ \and
        Nong Sang$^{1}$
}

\authorrunning{Changqian Yu et al.} %

\institute{Changqian Yu, Changxin Gao, Nong Sang \at
              \email{\{changqian\_yu, cgao, nsang\}@hust.edu.cn}
          \and
          Chunhua Shen \at
          	  \email{chunhua.shen@adelaide.edu.au}
          \and
          $^1$National Key Laboratory of Science and Technology on Multispectral Information Processing, School of Artificial Intelligence and Automation, Huazhong University of Science and Technology, Wuhan, China\\
          $^2$%
          The University of Adelaide,  Australia\\
          $^3$The Chinese University of Hong Kong\\
          $^4$Tencent\\
          $^*$Corresponding author
}

\date{Received: date / Accepted: date}

\maketitle
\begin{abstract}
The low-level details and high-level semantics are both essential to the semantic segmentation task.
However, to 
speed up 
the model inference, 
current appr\-oa\-ch\-es 
almost 
always sacrifice the low-level details, which leads to a 
considerable 
accuracy decrease.
We
propose to
treat these spatial details and categorical semantics separately to achieve high accuracy and high efficiency for real-time semantic segmentation.
To this end, we propose an efficient and effective architecture with a good trade-off between speed and accuracy, termed Bilateral Segmentation Network (BiSeNet V2).
This architecture involves: (\romannumeral1) a Detail Branch, with wide channels and shallow layers to capture low-level details and generate high-resolution feature representation;
(\romannumeral2) a Semantic Branch, with narrow channels and deep layers to obtain high-level semantic context.
The Semantic Branch is lightweight due to reducing the channel capacity and a fast-downsampling strategy.
Furthermore, we design a Guided Aggregation Layer to enhance mutual connections and fuse both types of feature representation.
Besides, a booster training strategy is 
designed 
to improve the segmentation performance without any extra inference cost.
Extensive quantitative and qualitative evaluations demonstrate that the proposed architecture performs favourably against 
a few 
\emph{state-of-the-art} real-time semantic segmentation appr\-oa\-ch\-es.
Specifically, for a 2,048$\times$1,024 input, we achieve 72.6\% Mean IoU on the Cityscapes test set with a speed of 156 FPS on one NVIDIA GeForce GTX 1080 Ti card, which is significantly faster than  existing methods,
yet we achieve better segmentation accuracy. 
Code and trained models will be made publicly available.
\end{abstract}

\begin{figure}[t]
\footnotesize
\centering
\renewcommand{\tabcolsep}{1pt} %
\ra{1} %
\begin{center}
\includegraphics[width=\linewidth]{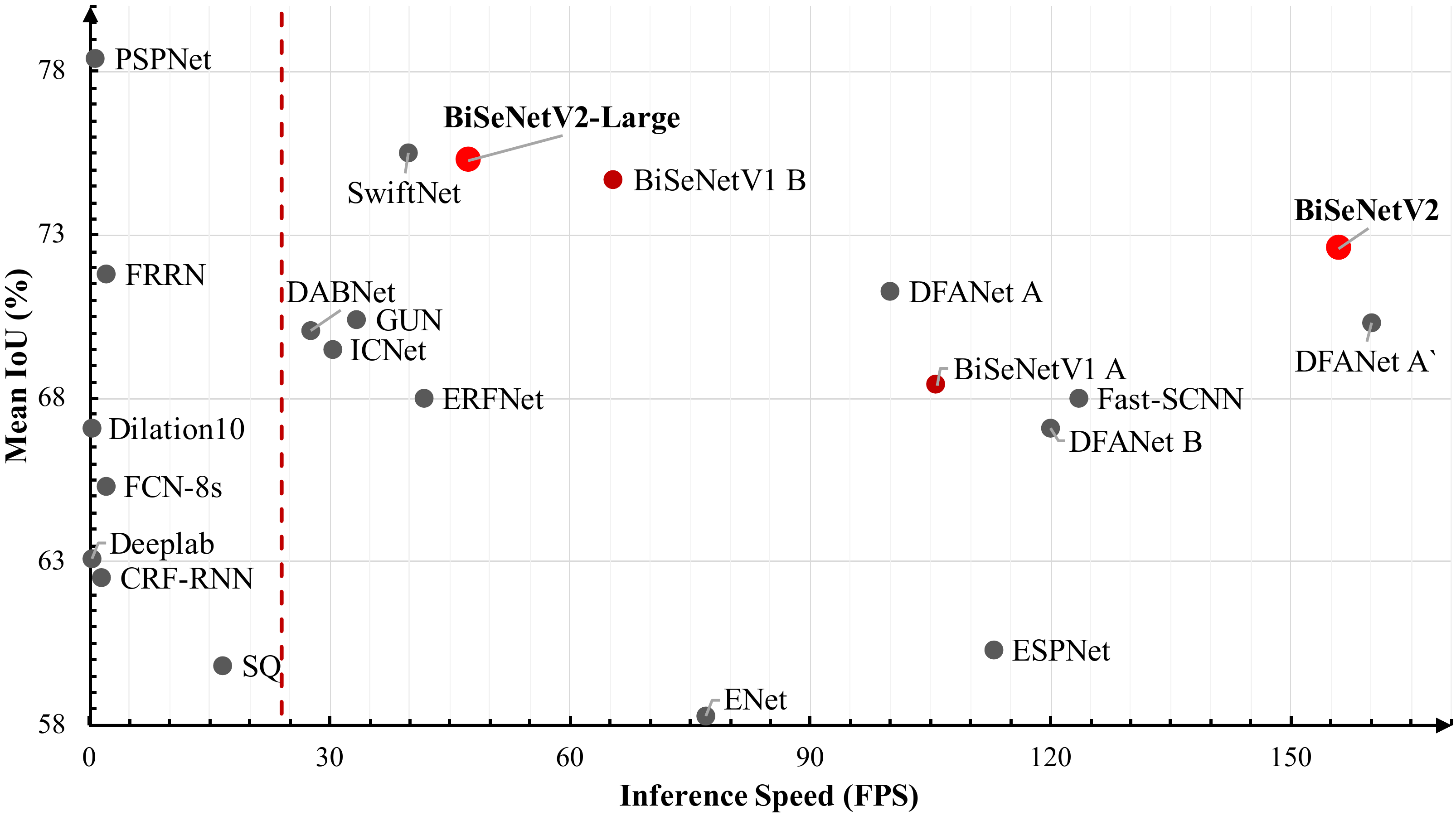}
\end{center}
\caption{\textbf{Speed-accuracy trade-off comparison on the Cityscapes \emph{test} set}.
Red dots indicate our methods, while grey dots means other methods.
The red line represents the real-time speed. 
}
\label{fig:comparison}
\vspace{-1em}
\end{figure}

\section{Introduction}
\label{sec:intro}
Semantic segmentation is the task of assigning semantic labels to each pixel.
It is a fundamental problem in computer vision with extensive applications, 
including 
scene understanding~\citep{Zhou-ADE-2016}, autonomous driving \citep{Cityscapes, Kitti}, human-machine interaction and video surveillance, just to name a few.
In recent years, with the advance of convolutional neural network~\citep{Krizhevsky-NIPS-Imagenet}, a series of semantic segmentation methods \citep{Zhao-CVPR-PSPNet-2017, Chen-Arxiv-Deeplabv3-2017, Yu-CVPR-DFN-2018, Chen-ECCV-Deeplabv3p-2018, Zhang-CVPR-EncNet-2018} based on fully convolutional network (FCN) \citep{Long-CVPR-FCN-2015} have constantly 
advanced the
{state-of-the-art} performance.

\begin{figure*}[t]
\footnotesize
\centering
\renewcommand{\tabcolsep}{1pt} %
\renewcommand{\arraystretch}{1} %
\begin{center}
\begin{tabular}{ccc}
\includegraphics[width=0.3\linewidth]{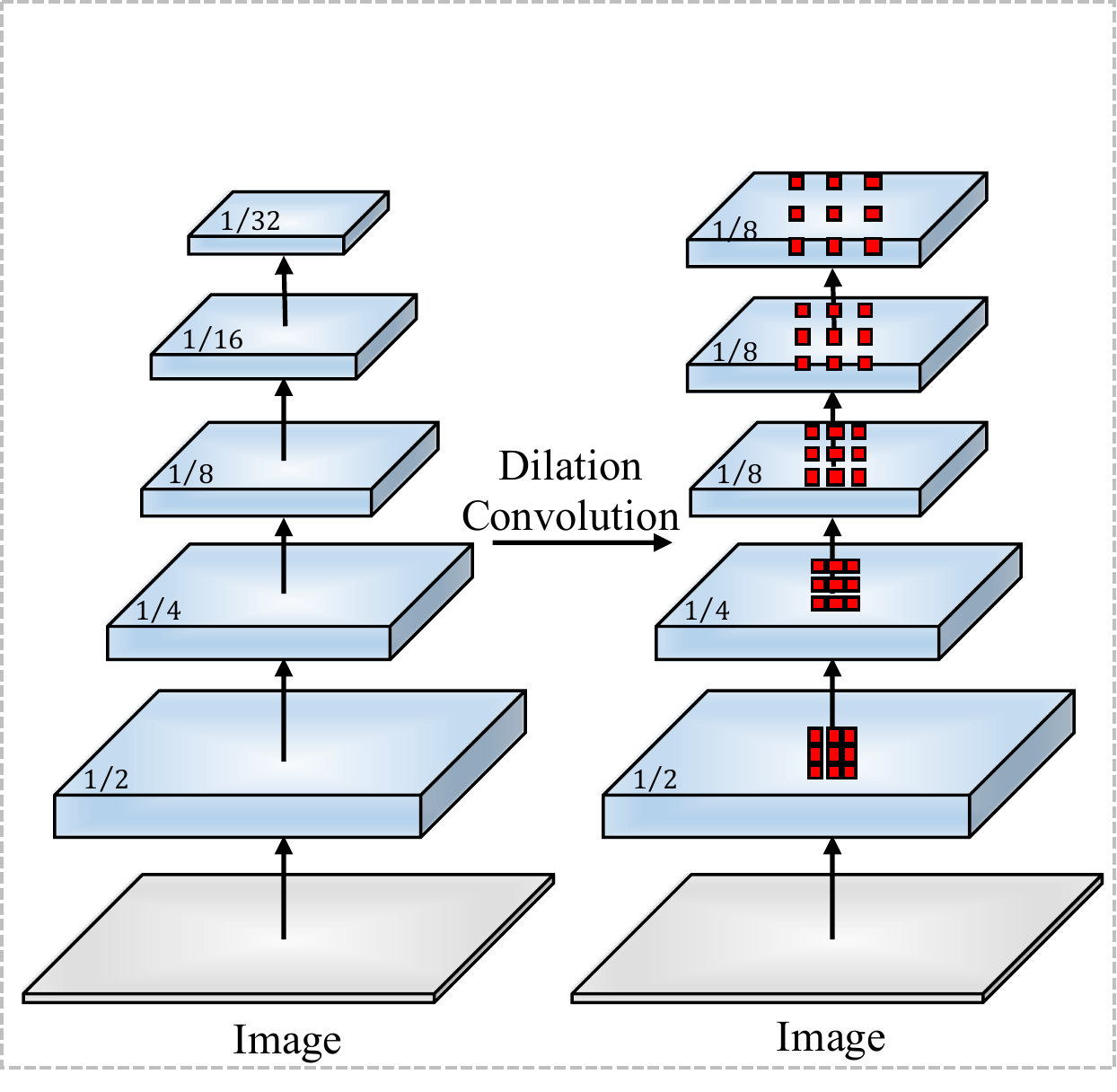} &
\includegraphics[width=0.3\linewidth]{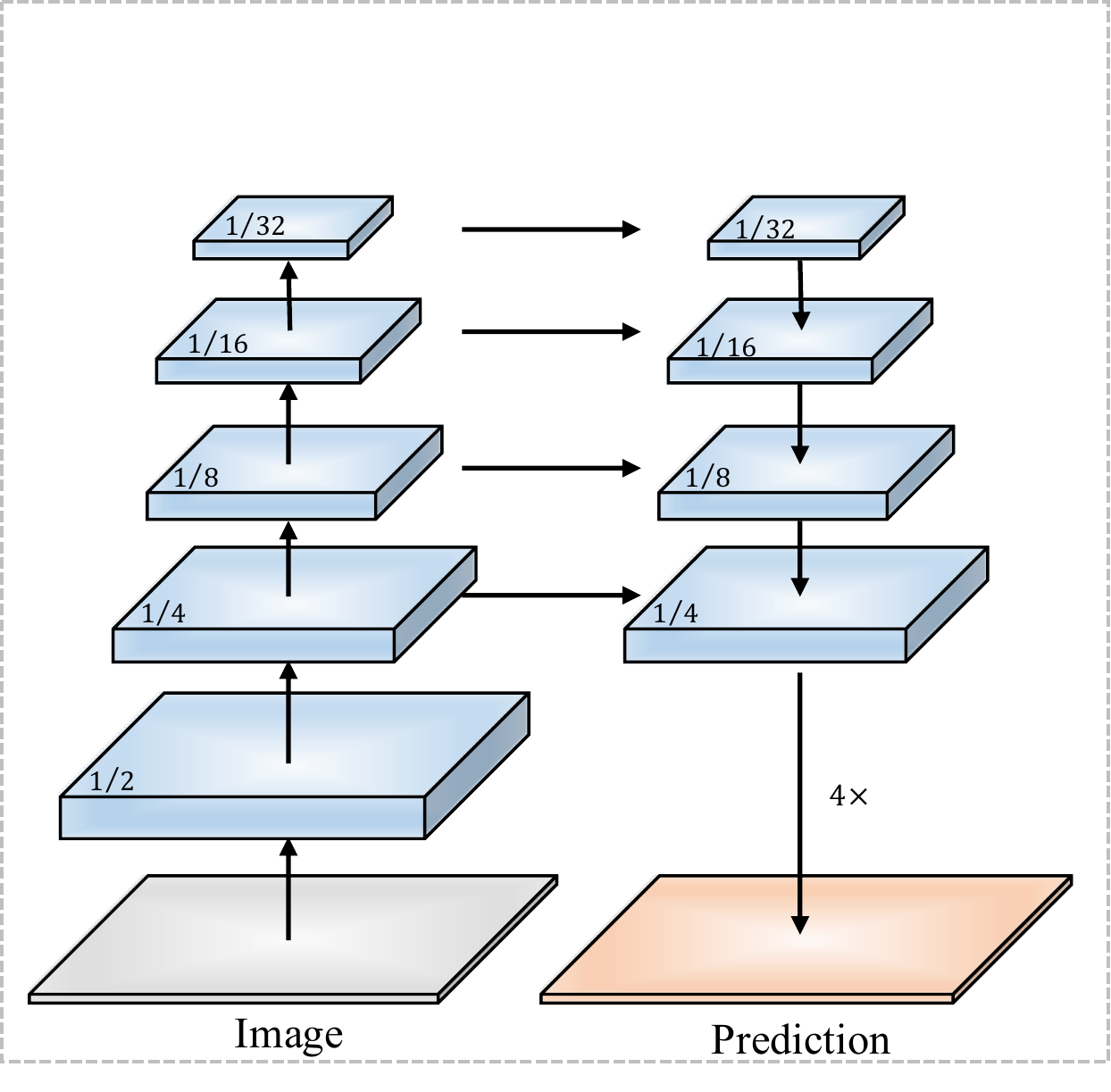} &
\includegraphics[width=0.3\linewidth]{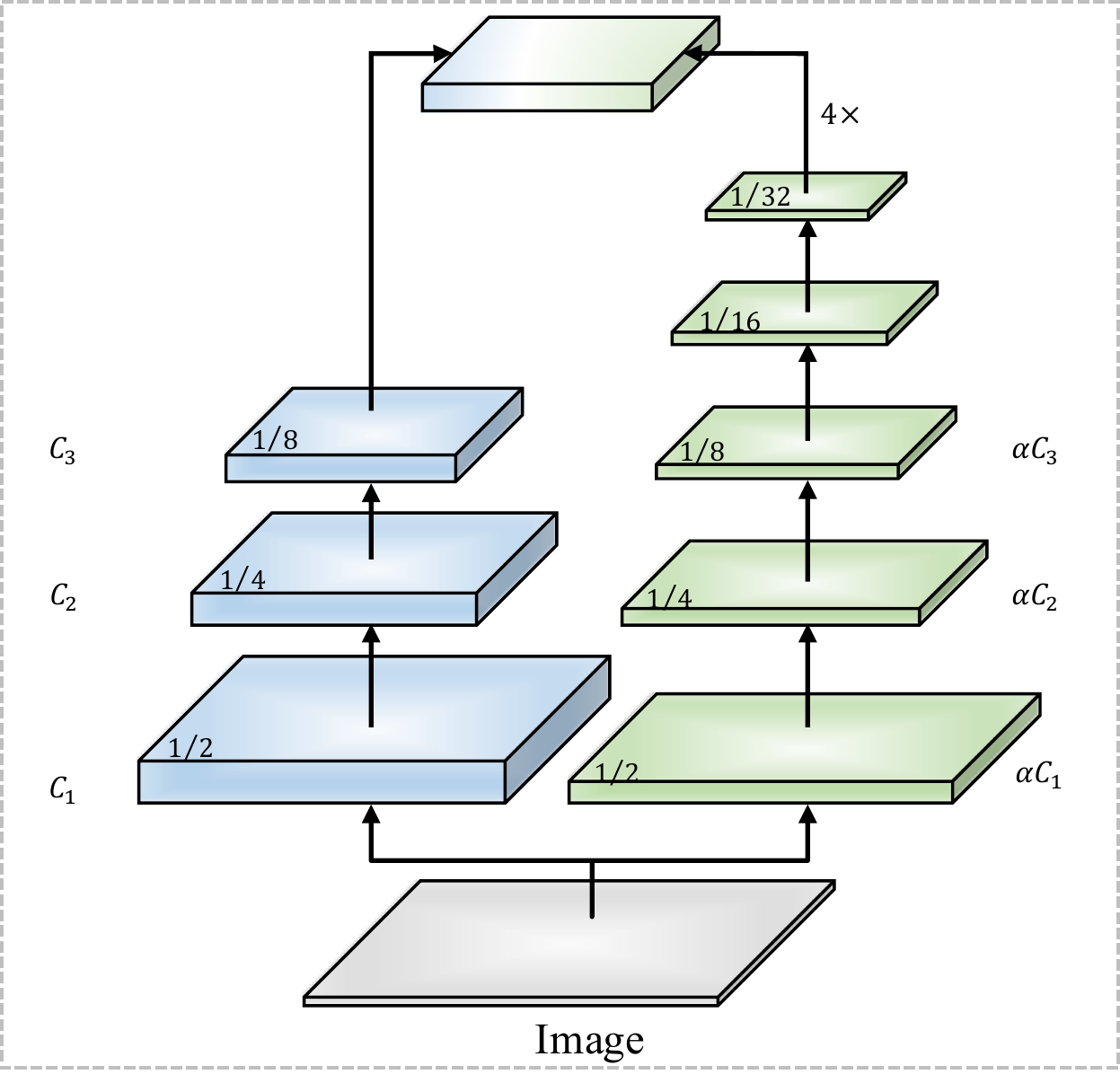} \\
(a) Dilation Backbone &
(b) Encoder-Decoder Backbone &
(c) Bilateral Segmentation Backbone \\
\end{tabular}
\end{center}
\caption{\textbf{Illustration of different backbone architectures.} 
(a) is the dilation backbone network, which removes the downsampling operations and upsampling the corresponding convolution filters.
It has heavy computation complexity and memory footprint.
(b) is the encoder-decoder backbone network, which adds extra top-down and lateral connections to recover the high-resolution feature map.
These connections in the network are less friendly to the memory access cost.
To achieve high accuracy and high efficiency simultaneously, we design the (c) Bilateral Segmentation backbone network.
This architecture has two pathways, Detail Branch for spatial details and Semantic Branch for categorical semantics.
The detail branch has wide channels and shallow layers, while the semantic branch has narrow channels and deep layers, which can be made very lightweight by the factor ($\lambda$, \eg $1/4$).
}
\label{fig:fig1}
\end{figure*}

The high accuracy of these methods depends on their backbone networks.
There are two main architectures as the backbone networks: (\romannumeral1) \textit{Dilation Backbone}, 
removing the downsampling operations and upsampling the corresponding filter kernels to maintain high-resolution feature representation~\citep{Chen-ICLR-Deeplabv2-2016, Chen-ECCV-Deeplabv3p-2018, Zhao-CVPR-PSPNet-2017, Zhao-ECCV-PSANet-2018, Fu-CVPR-DANet-2019, Yu-CVPR-CPN-2020}, as shown in Figure~\ref{fig:fig1} (a).
(\romannumeral2) \textit{Encoder-Decoder Backbone},
with top-down and skip connections to recover the high-resolution feature representation in the decoder part~\citep{Lin-CVPR-Refinenet-2017, Peng-CVPR-Largekernl-2017, Yu-CVPR-DFN-2018}, as illustrated in Figure~\ref{fig:fig1} (b).
However, both architectures are designed for general semantic segmentation tasks  with less 
care about 
the inference speed and computational cost.
In the dilation backbone, the dilation convolution is time-consuming and removing down-sampling operation brings heavy computation complexity and memory footprint.
Numerous connections in the encoder-decoder architecture are less friendly to the memory access cost~\citep{Ma-ECCV-Shufflenetv2-2018}.
However, the real-time semantic segmentation applications 
demand for an efficient inference speed.

Facing %
this demand, based on both backbone networks, existing methods \citep{Badrinarayanan-PAMI-SegNet-2017, Paszke-Arxiv-ENet-2016, Zhao-ECCV-ICNet-2018, Romera-TITS-ERFNet-2018, Mazzini-BMVC-GUN-2018} mainly 
employ 
two appraches to accelerate the model: (\romannumeral1) \textit{Input Restricting}. 
Smaller input resolution results in less computation cost with the same network architecture. 
To achieve real-time inference sp\-eed, 
many 
algorithms \citep{Zhao-ECCV-ICNet-2018, Romera-TITS-ERFNet-2018, Mazzini-BMVC-GUN-2018, Romera-TITS-ERFNet-2018} attempt to restrict the input size to reduce the whole computation complexity;
(\romannumeral2)\textit{Channel Pruning}.
It is a straight-forward acceleration method, especially pruning channels in early stages to boost inference speed \citep{Badrinarayanan-PAMI-SegNet-2017, Paszke-Arxiv-ENet-2016, Chollet-CVPR-Xception-2017}.
Although both manners can improve the inference speed to some extent, they sacrifice the low-level details and spatial capacity leading to a dramatic accuracy decrease.
Therefore, to achieve high efficiency and high accuracy simultaneously, it is challenging and of great importance to exploit a specific architecture for the real-time semantic segmentation task.

We observe that both of the low-level details and high-level semantics are crucial to the semantic segmentation task.
In the general semantic segmentation task, the deep and wide networks encode both information simultaneously.
However, in the real-time semantic segmentation task, we can treat spatial details and categorical semantics separately to achieve the trade-off between the accuracy and inference speed.

To this end, we propose a two-pathway architecture, termed \textbf{Bi}lateral \textbf{Se}gmentation \textbf{Net}work (BiSeNet V2), for real-time semantic segmentation.
One pathway is designed to capture the spatial details with wide channels and shallow layers, called \textit{Detail Branch}.
In contrast, the other pathway is introduced to extract the categorical semantics with narrow channels and deep layers, called \textit{Semantic Branch}.
The Semantic Branch simply requires a large receptive field to capture semantic context, while the detail information can be supplied by the Detail Branch.
Therefore, the Semantic Branch can be made very lightweight with fewer channels and a fast-downsampling strategy.
Both types of feature representation are merged to construct a stronger and more comprehensive feature representation.
This conceptual design leads to an efficient and effective architecture for real-time semantic segmentation, as illustrated in Figure~\ref{fig:fig1} (c).

Specifically, in this study, we design a \textit{Guided Aggregation Layer} to merge both types of features effectively.
To further improve the performance without increasing the inference complexity, we present a \textit{booster} training strategy with a series of auxiliary prediction heads, which can be discarded in the inference phase.
Extensive quantitative and qualitative evaluations demonstrate that the proposed architecture performs favourably against \emph{state-of-the-art} real-time semantic segmentation approaches, as shown in Figure~\ref{fig:comparison}.

The main contributions are summarized as follows:
\begin{itemize}
\item We propose an efficient and effective two-pathway architecture, termed  Bilateral Segmentation Network, for real-time semantic segmentation, which treats the spatial details and categorical semantics separately. 
\item For the Semantic Branch, we design a new light-weight network based on depth-wise convolutions to enhance the receptive field and capture rich contextual information.
\item A booster training strategy is introduced to further improve the segmentation performance without increasing the inference cost.
\item Our architecture achieves impressive results on different benchmarks of Cityscapes~\citep{Cityscapes}, CamVid \citep{CamVid}, and COCO-Stuff \citep{COCO-Stuff}. 
More specifically, we obtain the results of $72.6\%$ mean IoU on the Cityscapes \emph{test} set with the speed of $156$ FPS on one NVIDIA GeForce GTX 1080Ti card.
\end{itemize}

A
preliminary version of this work 
was 
published in \citep{Yu-ECCV-BiSeNet-2018}. 
We have 
extended
our conference version as follows.  
(\romannumeral1) We simplify the original structure to present an efficient and effective architecture for %
real-time semantic segmentation.
We remove the time-consuming cross-layer connections in the original version to obtain a more clear and simpler architecture.
(\romannumeral2) We re-design the overall architecture with more compact network structures and well-designed components.
Specifically, we deepen the Detail Path to encode more details.
We design 
light-weight components based on the depth-wise convolutions for the Semantic Path.
Meanwhile, we propose an efficient aggregation layer to enhance the mutual connections between both paths. 
(\romannumeral3) We conduct comprehensive ablative experiments to elaborate on the effectiveness and efficiency of the proposed method.
(\romannumeral4) We 
have significantly 
improved the accuracy and speed of the method in our previous work, \ie for a $2048\times1024$ input, achieving $72.6\%$ Mean IoU on the Cityscapes \emph{test} set with a speed of $156$ FPS on one NVIDIA GeForce GTX 1080Ti card. 

\section{Related Work}
\label{sec:related-work}
Recent years have witnessed significant advances in image semantic segmentation.
In this section, our discussion mainly focuses on three groups of methods most relevant to our work, \ie generic semantic segmentation methods, real-time semantic segmentation methods, and light-weight architectures.

\subsection{Generic Semantic Segmentation}
Traditional segmentation methods based on the threshold selection \citep{Otsu-TSMC-OTSU-1979}, the region growing \citep{Vincent-PAMI-Watersheds-1991}, the super-pixel \citep{Ren-ICCV-Superpixel-2003, Achanta-PAMI-SLIC-2012, Van-ECCV-Seeds-2012} and the graph \citep{Boykov-ICCV-GraphCut-2001, Rother-TOG-Grbcut-2004} algorithms adopt the hand-crafted features to solve this problem.
Recently, a new generation of algorithms based on FCN \citep{Long-CVPR-FCN-2015, Shelhamer-PAMI-FCN-2017} keep improving \emph{state-of-the-art} performance on different benchmarks.
Various methods are based on two types of backbone network: (\romannumeral1) \textit{dilation backbone network}; (\romannumeral2) \textit{encoder-decoder backbone network}.

On 
one hand, the dilation backbone removes the downsampling operations and upsamples the convolution filter to preserve high-resolution feature representations.
Due to the simplicity of the dilation convolution, various methods \citep{Chen-ICLR-Deeplabv2-2016, Chen-ECCV-Deeplabv3p-2018, Zhao-CVPR-PSPNet-2017, Wang-WACV-DUC-2018, Zhang-CVPR-EncNet-2018, Yu-CVPR-CPN-2020} develop different novel and effective components on it. 
The Deeplabv3~\citep{Chen-Arxiv-Deeplabv3-2017} devises an atrous spatial pyramid pooling to capture multi-scale context, while the PSPNet~\citep{Zhao-CVPR-PSPNet-2017} adopts a pyramid pooling module on the dilation backbone.
Meanwhile, some methods introduce the attention mechanisms, \eg self-attention~\citep{Yuan-Arxiv-OCNet-2018, Fu-CVPR-DANet-2019, Yu-CVPR-CPN-2020}, spatial attention~\citep{Zhao-ECCV-PSANet-2018} and channel attention~\citep{Zhang-CVPR-EncNet-2018},  to capture long-range context based on the dilation backbone.

On the other hand, the encoder-decoder backbone network adds extra top-down and lateral connections to recover the high-resolution feature maps in the decoder part.
FCN and Hypercolumns \citep{Hariharan-CVPR-HyperNet-2015} adopt the skip connection to integrate the low-level feature.
Meanwhile, U-net \citep{Ronneberger-MICCAI-U-net-2015}, SegNet with saved pooling indices \citep{Badrinarayanan-PAMI-SegNet-2017}, RefineNet with multi-path refinement \citep{Lin-CVPR-Refinenet-2017}, LRR with step-wise reconstruction \citep{Ghiasi-ECCV-LRR-2016}, GCN with ``large kernel'' convolution \citep{Peng-CVPR-Largekernl-2017} and DFN with channel attention module \citep{Yu-CVPR-DFN-2018} incorporate this backbone network to recover the detailed information.
HRNet \citep{Wang-TPAMI-HRNet-2019} adopts multi-branches to maintain the high resolution.

Both types of backbone network encode the low-level details and high-level semantics simultaneously with the wide and deep networks.
Although both types of backbone network achieve %
{state-of-the-art} perfor\-mance, most methods run at a slow inference speed.
In this study, we propose a novel and efficient architecture to treat the spatial details and categorical semantics separately to achieve a good trad-off between segmentation accuracy and inference speed.

\begin{figure*}[t]
\footnotesize
\centering
\renewcommand{\tabcolsep}{1pt} %
\renewcommand{\arraystretch}{1} %
\begin{center}
\includegraphics[width=0.98\linewidth]{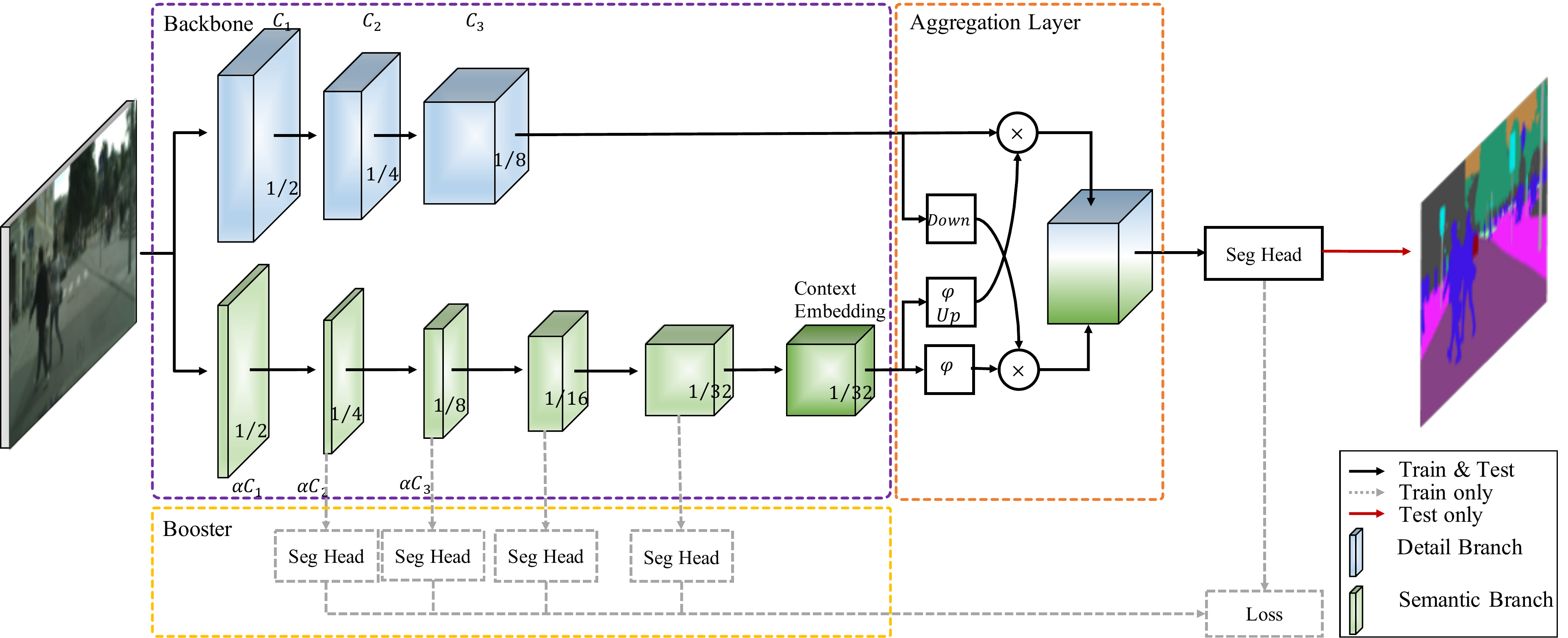}
\end{center}
\caption{\textbf{Overview of the Bilateral Segmentation Network.}
There are mainly three 
components: 
two-pathway backbone in the {\color{Purple}{purple}} dashed box, the aggregation layer in the {\color{orange}{orange}} dashed box, and the booster part in the {\color{Gold}{yellow}} dashed box.
The two-pathway backbone has a Detail Branch (the {\color{Cerulean}{blue}} cubes) and a Semantic Branch (the {\color{LightGreen}{green}} cubes).
The three stages in Detail Branch have $C_1, C_2, C_3$ channels respectively.
The channels of corresponding stages in Semantic Branch can be made lightweight by the factor $\lambda(\lambda<1)$. 
The last stage of the Semantic Branch is the output of the Context Embedding Block. 
Meanwhile, numbers in the cubes are the feature map size ratios to the resolution of the input.
In the Aggregation Layer part, we adopt the bilateral aggregation layer.
$Down$ indicates the downsampling operation, $Up$ represents the upsampling operation, $\varphi$ is the Sigmoid function, and $\bigotimes$ means element-wise product. 
Besides, in the booster part, we design some auxiliary segmentation heads to improve the segmentation performance without any extra inference cost.
}
\label{fig:network}
\end{figure*}

\subsection{Real-time Semantic Segmentation}
Real-time semantic segmentation algorithms attract increasing attention when a growing practical applications require fast interaction and response.
SegNet \citep{Badrinarayanan-PAMI-SegNet-2017} uses a small network structure and the skip connection to achieve a fast speed. 
E-Net \citep{Paszke-Arxiv-ENet-2016} devises a lightweight network from scratch and delivers extremely high speed. 
ICNet \citep{Zhao-ECCV-ICNet-2018} uses the image cascade to speed up the algorithm, while DLC \citep{Li-CVPR-DLC-2017} employs a cascade network structure to reduce the computation in ``easy regions''.
ERFNet \citep{Romera-TITS-ERFNet-2018} adopts the residual connection and factorized convolutions to remain efficient and retain accuracy.
Meanwhile, ESPNet \citep{Mehta-ECCV-ESPNet-2018, Mehta-CVPR-ESPNetV2-2019} devises an efficient spatial pyramid dilated convolution for real-time semantic segmentation.
GUN \citep{Mazzini-BMVC-GUN-2018} employs a guided upsampling module to fuse the information of the multi-resolution input.
DFANet \citep{Li-CVPR-DFANet-2019} reuses the feature to enhance the feature representation and reduces the complexity.

Although these methods can achieve a real-time inference speed, they dramatically sacrifice the accuracy to the efficiency with the loss of the low-level details. 
In this work, we take both of the low-level details and high-level semantics into consideration to achieve high accuracy and high efficiency.

\subsection{Light-weight Architecture}
Following the pioneering work of group/depth-wise convolution and separable convolution, light-weight architecture design has achieved rapid development, including Xception \citep{Chollet-CVPR-Xception-2017}, MobileNet \citep{Howard-Arxiv-MobileNet-2017, Sandler-CVPR-MobileNetv2-2018}, ShuffleNet \citep{Zhang-CVPR-Shufflenet-2018, Ma-ECCV-Shufflenetv2-2018}, to name a few.
These methods achieve a valuable trade-off between speed and accuracy for the classification task.
In this study, we design a light-weight network given computation complexity, memory access cost and real inference speed for the real-time semantic segmentation.

\section{Core Concepts 
of 
BiSeNetV2}
\label{sec:concepts}
Our architecture 
consists of 
a Detail Branch (Section~\ref{sec:concept:detail-branch}) and a Semantic Branch (Section~\ref{sec:concept:semantic-branch}), which are merged by an Aggregation Layer (Section~\ref{sec:concept:aggregation-layer}).
In this section, we demonstrate the core concepts of our architecture, as illustrated in Figure~\ref{fig:fig1}(c).

\begin{table*}[t]
\scriptsize
\centering
\ra{1.2}
\caption{\textbf{Instantiation of the Detail Branch and Semantic Branch.}
Each stage $S$ contains one or more operations $opr$ (\eg \textit{Conv2d}, \textit{Stem}, \textit{GE}, \textit{CE}).
Each operation has a kernels size $k$, stride $s$ and output channels $c$, repeated $r$ times.
The expansion factor $e$ is applied to expand the channel number of the operation.
Here the channel ratio is $\lambda=1/4$. The green colors mark \textit{fewer} channels of Semantic Branch in the corresponding stage of the Detail Branch.
Notation:
\textit{Conv2d} means the convolutional layer, followed by one batch normalization layer and relu activation function.
\textit{Stem} indicates the stem block.
\textit{GE} represents the gather-and-expansion layer.
\textit{CE} is the context embedding block.
}
\label{tab:detail-and-semantic-branch}
\begin{tabular}{c|lcccc|lccccc|l}
\shline
\multirow{2}{*}{Stage} & \multicolumn{5}{c|}{Detail Branch}           & \multicolumn{6}{c|}{Semantic Branch}                                                                                        & Output Size     \\ \cline{2-13} 
                       & $opr$                & $k$ & $c$ & $s$ & $r$ & $opr$   & $k$                  & $c$                  & $e$                  & $s$                  & $r$                   &                 \\ \hline
Input                  &                      &     &     &     &     &         &                      &                      &                      &                      &                       & 512$\times$1024 \\ \hline
\multirow{2}{*}{$S_1$} & Conv2d               & 3   & 64  & 2   & 1   & Stem    & 3                    & \green{16}           & -                    & 4                    & 1                     & 256$\times$512  \\
                       & Conv2d               & 3   & 64  & 1   & 1   &         & \multicolumn{1}{l}{} & \multicolumn{1}{l}{} & \multicolumn{1}{l}{} & \multicolumn{1}{l}{} & \multicolumn{1}{l|}{} & 256$\times$512  \\ \cline{1-6} \cline{13-13} 
\multirow{2}{*}{$S_2$} & Conv2d               & 3   & 64  & 2   & 1   &         &                      &                      &                      &                      &                       & 128$\times$256  \\
                       & Conv2d               & 3   & 64  & 1   & 2   &         &                      &                      &                      &                      &                       & 128$\times$256  \\ \hline
\multirow{2}{*}{$S_3$} & Conv2d               & 3   & 128 & 2   & 1   & GE  & 3                    & \green{32}           & 6                    & 2                    & 1                     & 64$\times$128   \\
                       & Conv2d               & 3   & 128 & 1   & 2   & GE & 3                    & \green{32}           & 6                    & 1                    & 1                     & 64$\times$128   \\ \hline
\multirow{2}{*}{$S_4$} &                      &     &     &     &     & GE  & 3                    & 64                   & 6                    & 2                    & 1                     & 32$\times$64    \\
                       &                      &     &     &     &     & GE & 3                    & 64                   & 6                    & 1                    & 1                     & 32$\times$64    \\ \hline
\multirow{3}{*}{$S_5$} &                      &     &     &     &     & GE  & 3                    & 128                  & 6                    & 2                    & 1                     & 16$\times$32    \\
                       &                      &     &     &     &     & GE & 3                    & 128                  & 6                    & 1                    & 3                     & 16$\times$32    \\
                       & \multicolumn{1}{c}{} &     &     &     &     & CE & 3                    & 128                  & -                    & 1                    & 1                     & 16$\times$32    \\ \shline
\end{tabular}
\end{table*}

\subsection{Detail Branch}
\label{sec:concept:detail-branch}
The Detail Branch is responsible for the spatial details, which is low-level information.
Therefore, this branch requires a rich channel capacity to encode affluent spatial detailed information.
Meanwhile, because the Detail Branch simply focuses on the low-level details, we can design a shallow structure with a \textit{small} stride for this branch.
Overall, the key concept of the Detail Branch is to use \textit{wide} channels and \textit{shallow} layers for the spatial details.
Besides, the feature representation in this branch has a large spatial size and wide channels.
Therefore, it is better not to adopt the residual connections, which increases the memory access cost and reduce the speed.

\subsection{Semantic Branch}
\label{sec:concept:semantic-branch}
In parallel to the Detail Branch, the Semantic Branch is designed to capture high-level semantics.
This branch has \textit{low} channel capacity, while the spatial details can be provided by the Detail Branch.
In contrast, in our experiments, the Semantic Branch has a ratio of $\lambda(\lambda<1)$ channels of the Detail Branch, which makes this branch lightweight.
Actually, the Semantic Branch can be any lightweight convolutional model (\eg \citep{Chollet-CVPR-Xception-2017, Iandola-Arxiv-SqueezeNet-2016, Howard-Arxiv-MobileNet-2017, Sandler-CVPR-MobileNetv2-2018, Zhang-CVPR-Shufflenet-2018, Ma-ECCV-Shufflenetv2-2018}).
Meanwhile, the Semantic Branch adopts the fast-downsampling strategy to promote the level of the feature representation and enlarge the receptive field quickly.
High-level semantics require \textit{large} receptive field.
Therefore, the Semantic Branch employs the global average pooling~\citep{Liu-ARXIV-ParseNet-2016} to embed the global contextual response.

\subsection{Aggregation Layer}
\label{sec:concept:aggregation-layer}
The feature representation of the Detail Branch and the Semantic Branch is complementary, one of which is unaware of the information of the other one.
Thus, 
an Aggregation Layer is designed to merge both types of feature representation.
Due to the fast-downsampling strategy, the spatial dimensions of the Semantic Branch's output are smaller than the Detail Branch.
We need to upsample the output feature map of the Semantic Branch to match the output of the Detail Branch.

There are 
a few 
manners to fuse information, \eg simple \textit{summation}, \textit{concatenation} and some well-designed operations.
We have experimented different fusion methods with consideration of accuracy and efficiency.
At last, we adopt the bidirectional aggregation method, as shown in Figure~\ref{fig:network}.

\section{Bilateral Segmentation Network}
\label{sec:instantiations}
The concept of our BiSeNet is generic, which can be implemented with different convolutional models~\citep{He-CVPR-ResNet-2016, Huang-CVPR-DenseNet-2017, Chollet-CVPR-Xception-2017, Iandola-Arxiv-SqueezeNet-2016, Howard-Arxiv-MobileNet-2017, Sandler-CVPR-MobileNetv2-2018, Zhang-CVPR-Shufflenet-2018, Ma-ECCV-Shufflenetv2-2018} and any specific designs.
There are mainly three key concepts: 
(\romannumeral1) The Detail Branch has \textit{high} channel capacity and \textit{shallow} layers with \textit{small} receptive field for the spatial details; 
(\romannumeral2)The Semantic Branch has \textit{low} channel capacity and \textit{deep} layers with \textit{large} receptive field for the categorical semantics.
(\romannumeral3)An efficient Aggregation Layer is designed to fuse both types of representation.

In this subsection, according to the proposed conceptual design, we demonstrate our instantiations of the overall architecture and some other specific designs, as illustrated in Figure~\ref{fig:network}.

\subsection{Detail Branch}
\label{sec:instantiation:detail-branch}
The instantiation of the Detail Branch in Table~\ref{tab:detail-and-semantic-branch} contains three stages, each layer of which is a convolution layer followed by batch normalization~\citep{Ioffe-ICML-BN-2015} and activation function~\citep{Glorot-ICAIS-ReLU-2011}.
The first layer of each stage has a stride $s=2$, while the other layers in the same stage have the same number of filters and output feature map size.
Therefore, this branch extracts the output feature maps that are $1/8$ of the original input. 
This Detail Branch encodes rich spatial details due to the high channel capacity.
Because of the high channel capacity and the large spatial dimension, the residual structure~\citep{He-CVPR-ResNet-2016} will increases the memory access cost~\citep{Ma-ECCV-Shufflenetv2-2018}.
Therefore, this branch mainly obeys the philosophy of VGG nets~\citep{Simonyan-ICLR-VGG-2015} to stack the layers.

\begin{figure}[t]
\footnotesize
\centering
\renewcommand{\tabcolsep}{1pt} %
\renewcommand{\arraystretch}{1} %
\begin{center}
\begin{tabular}{cc}
\includegraphics[width=0.57\linewidth]{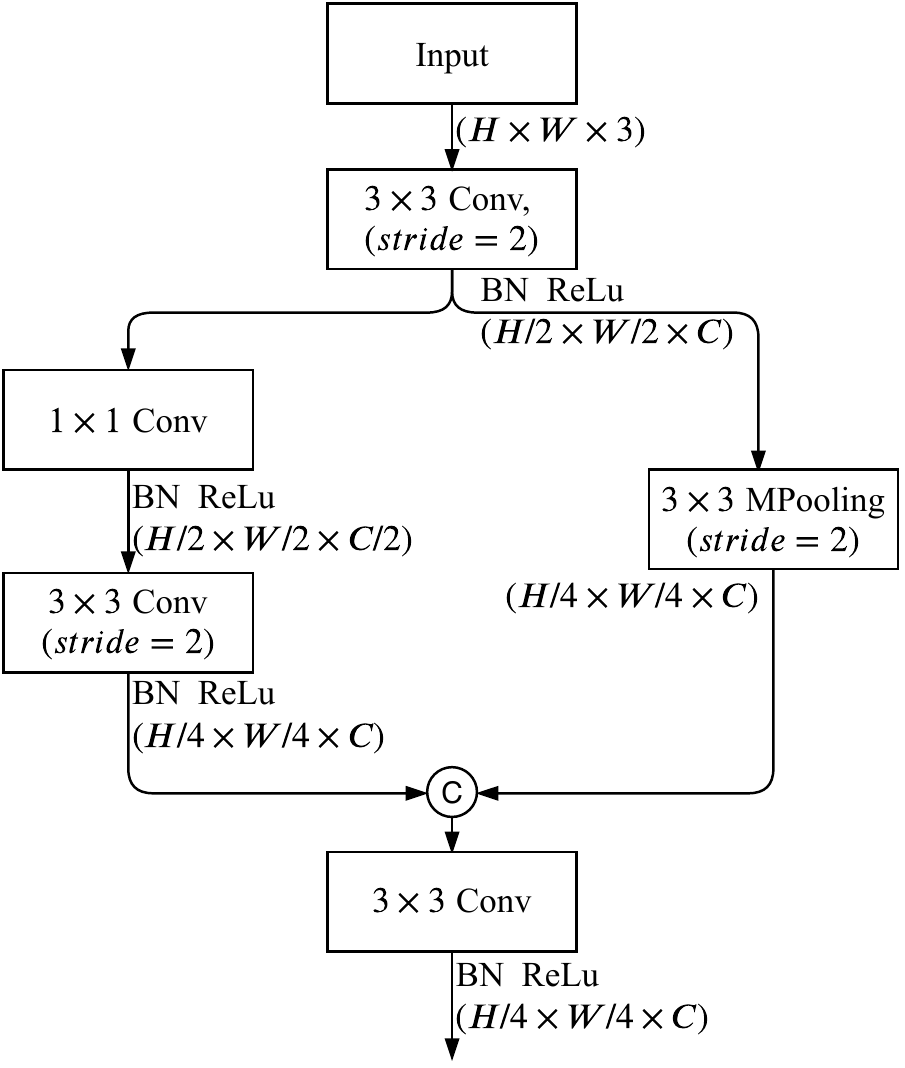} &
\includegraphics[width=0.35\linewidth]{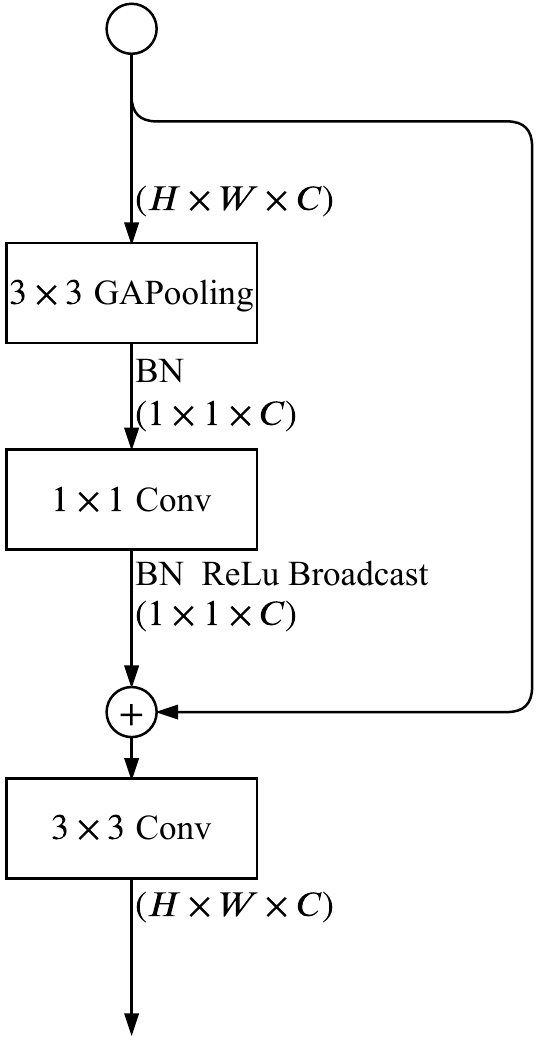} \\
(a) Stem Block & 
(b) Context Embedding Block \\
\end{tabular}
\end{center}
\caption{\textbf{Illustration of Stem Block and Context Embedding Block.}
(a) is the Stem Block, which adopts a fast-downsampling strategy.
This block has two branches with different manners to downsample the feature representation.
Then both feature response of two branches is concatenated as the output.
(b) is the Context Embedding Block.
As demonstrated in Section~\ref{sec:concept:semantic-branch}, the Semantic Branch requires large receptive field. 
Therefore, we design a Context Embedding Block with the global average pooling to embed the global contextual information.
Notation: \textit{Conv} is convolutional operation. \textit{BN} is the batch normalization. \textit{ReLu} is the ReLu activation function. \textit{Mpooling} is the max pooling. \textit{GPooling} is the global average pooling. and \textit{C} means concatenation.
Meanwhile, $1\times1, 3\times3$ denote the kernel size, $H\times W\times C$ means the tensor shape (height, width, depth).
}
\label{fig:stem-context-block}
\end{figure}

\subsection{Semantic Branch} 
\label{sec:instantiation:semantic-branch}
In consideration of the large receptive field and efficient computation simultaneously, we design the Semantic Branch, inspired by the philosophy of the lightweight recognition model, \eg Xception~\citep{Chollet-CVPR-Xception-2017}, MobileNet~\citep{Howard-Arxiv-MobileNet-2017, Sandler-CVPR-MobileNetv2-2018, Howard-ICCV-Mobilenetv3-2019}, ShuffleNet~\citep{Zhang-CVPR-Shufflenet-2018, Ma-ECCV-Shufflenetv2-2018}.
Some of the key features of the Semantic Branch are as follows.

\begin{figure}[t]
\footnotesize
\centering
\renewcommand{\tabcolsep}{1pt} %
\ra{1} %
\begin{center}
\begin{tabular}{ccc}
\includegraphics[width=0.3\linewidth]{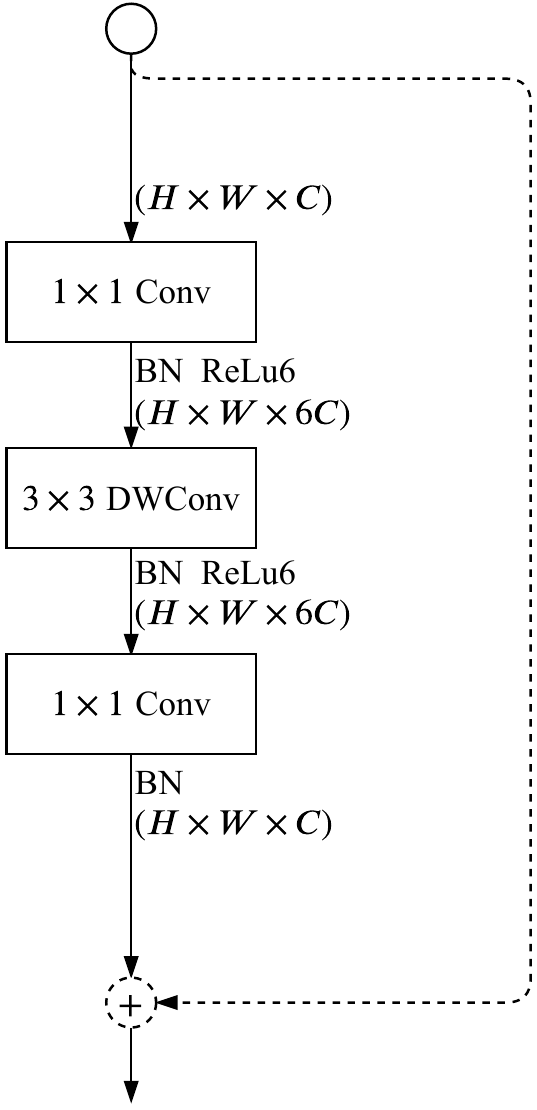} & 
\includegraphics[width=0.3\linewidth]{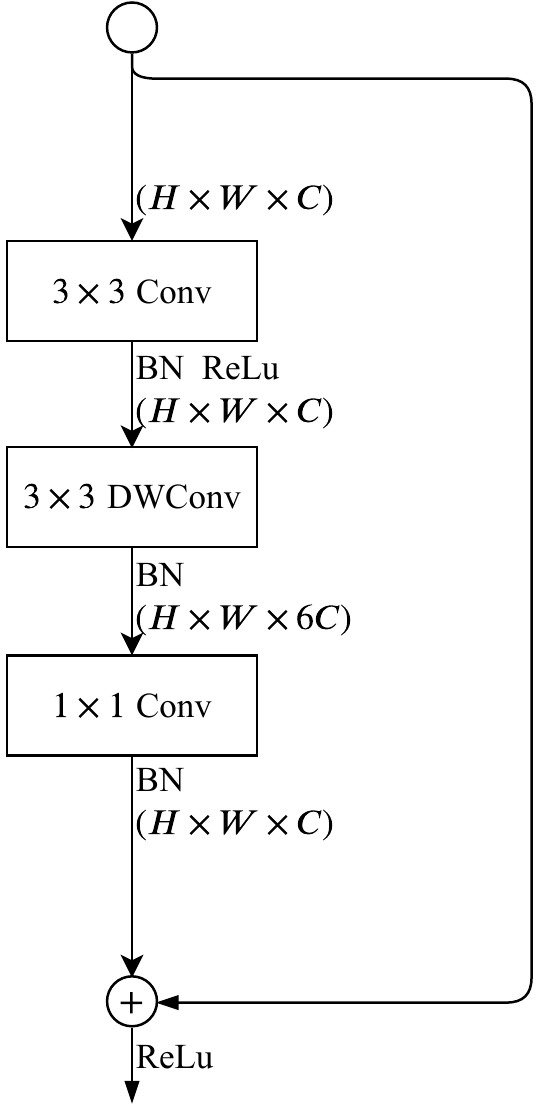} &
\includegraphics[width=0.37\linewidth]{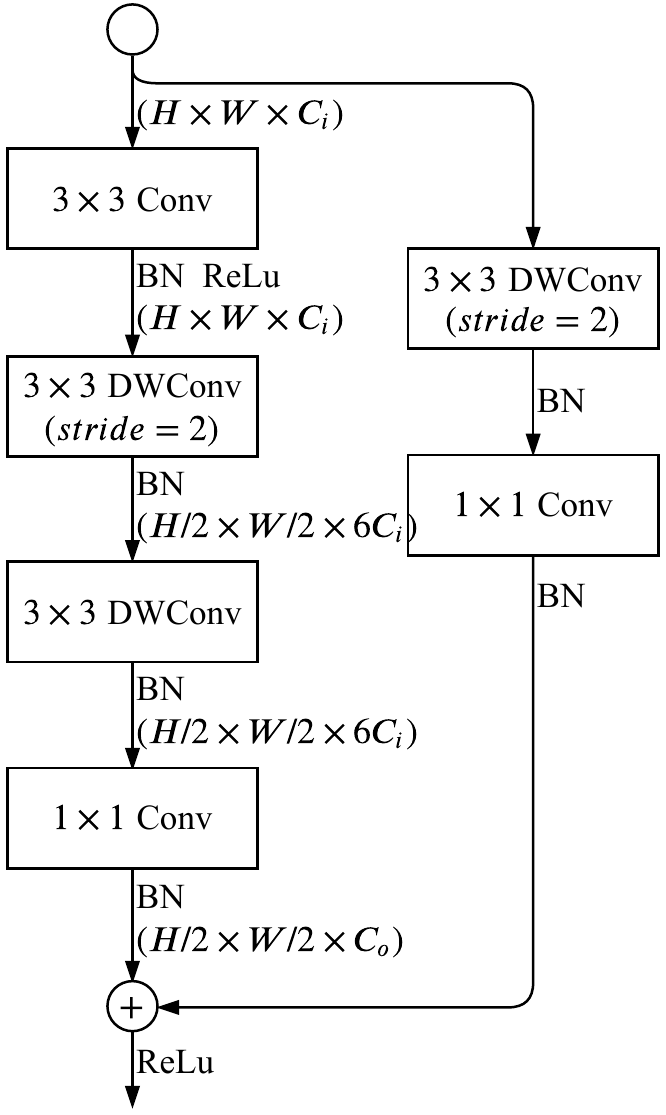} \\
(a) & (b) & (c) \\
\end{tabular}
\end{center}
\caption{\textbf{Illustration of Inverted Bottleneck and Gather-and-Expansion Layer}.
(a) is the mobile inverted bottleneck Conv proposed in MobileNetv2.
The dashed shortcut path and summation circle do not exist with the $stride=2$.
(b)(c) are the proposed Gather-and-Expansion Layer.
The bottleneck structure adopts: (\romannumeral1) a  $3\times3$ convolution to gather local feature response and expand to higher-dimensional space;(\romannumeral2) a $3\times3$ depth-wise convolution performed independently over each individual output channel of the expansion layer; (\romannumeral3) a $1\times1$ convolution as the projection layer to project the output of depth-wise convolution into a low channel capacity space.
When the $stride=2$, we adopt two $kernel\_size=3$ depth-wise convolutions on the main path and a $3\times3$ separable convolution as the shortcut.
Notation: \textit{Conv} is convolutional operation. \textit{BN} is the batch normalization. \textit{ReLu} is the ReLu activation function.
Meanwhile, $1\times1, 3\times3$ denote the kernel size, $H\times W\times C$ means the tensor shape (height, width, depth).
}
\label{fig:gelayer}
\end{figure}

\paragraph{\rm \textbf{Stem Block}} Inspired from~\citep{Szegedy-AAAI-Inceptionv4-2017, Shen-ICCV-DSOD-2017, Wang-NIPS-Pelee-2018}, we adopt the Stem Block as the first stage of the Semantic Branch, as illustrated in Figure~\ref{fig:stem-context-block}.
It uses two different downsampling manners to shrink the feature representation.
And then the output feature of both branches are concatenated as the output.
This structure has efficient computation cost and effective feature expression ability.

\paragraph{\rm \textbf{Context Embedding Block}} As discussed in Section~\ref{sec:concept:semantic-branch}, the Semantic Branch requires large receptive field to capture high-level semantics.
Inspired from~\citep{Yu-CVPR-DFN-2018, Liu-ARXIV-ParseNet-2016, Zhao-CVPR-PSPNet-2017, Chen-Arxiv-Deeplabv3-2017}, we design the Context Embedding Block.
This block uses the global average pooling and residual connection~\citep{He-CVPR-ResNet-2016} to embed the global contextual information efficiently, as showed in Figure~\ref{fig:stem-context-block}.

\paragraph{\rm \textbf{Gather-and-Expansion Layer}} Taking advantage of the benefit of depth-wise convolution, we propose the Gather-and-Expansion Layer, as illustrated in Figure~\ref{fig:gelayer}.
The Gather-and-Expansion Layer consists of: (\romannumeral1) a  $3\times3$ convolution to efficiently aggregate feature responses and expand to a higher-dimensional space; (\romannumeral3) a $3\times3$ depth-wise convolution performed independently over each individual output channel of the expansion layer; (\romannumeral4) a $1\times1$ convolution as the projection layer to project the output of depth-wise convolution into a low channel capacity space.
When $stide=2$, we adopt two $3\times3$ depth-wise convolution, which further enlarges the receptive field, and one $3\times3$ separable convolution as the shortcut.
Recent works~\citep{Tan-CVPR-MNASNet-2019, Howard-ICCV-Mobilenetv3-2019} adopt $5\times5$ separable convolution heavily to enlarge the receptive field, which has fewer FLOPS than two $3\times3$ separable convolution in some conditions.
In this layer, we replace the $5\times5$ depth-wise convolution in the separable convolution with two $3\times3$ depth-wise convolution, which has fewer FLOPS and the same receptive field.

In contrast to the inverted bottleneck in MobileNetv2, the GE Layer has one more $3\times3$ convolution.
However, this layer is also friendly to the computation cost and memory access cost~\citep{Ma-ECCV-Shufflenetv2-2018, Sandler-CVPR-MobileNetv2-2018}, because the $3\times3$ convolution is specially optimized in the CUDNN library~\citep{CUDNN, Ma-ECCV-Shufflenetv2-2018}. 
Meanwhile, because of this layer, the GE Layer has higher feature expression ability than the inverted bottleneck.

\begin{figure}[t]
\footnotesize
\centering
\renewcommand{\tabcolsep}{1pt} %
\ra{1} %
\begin{center}
\includegraphics[width=0.95\linewidth]{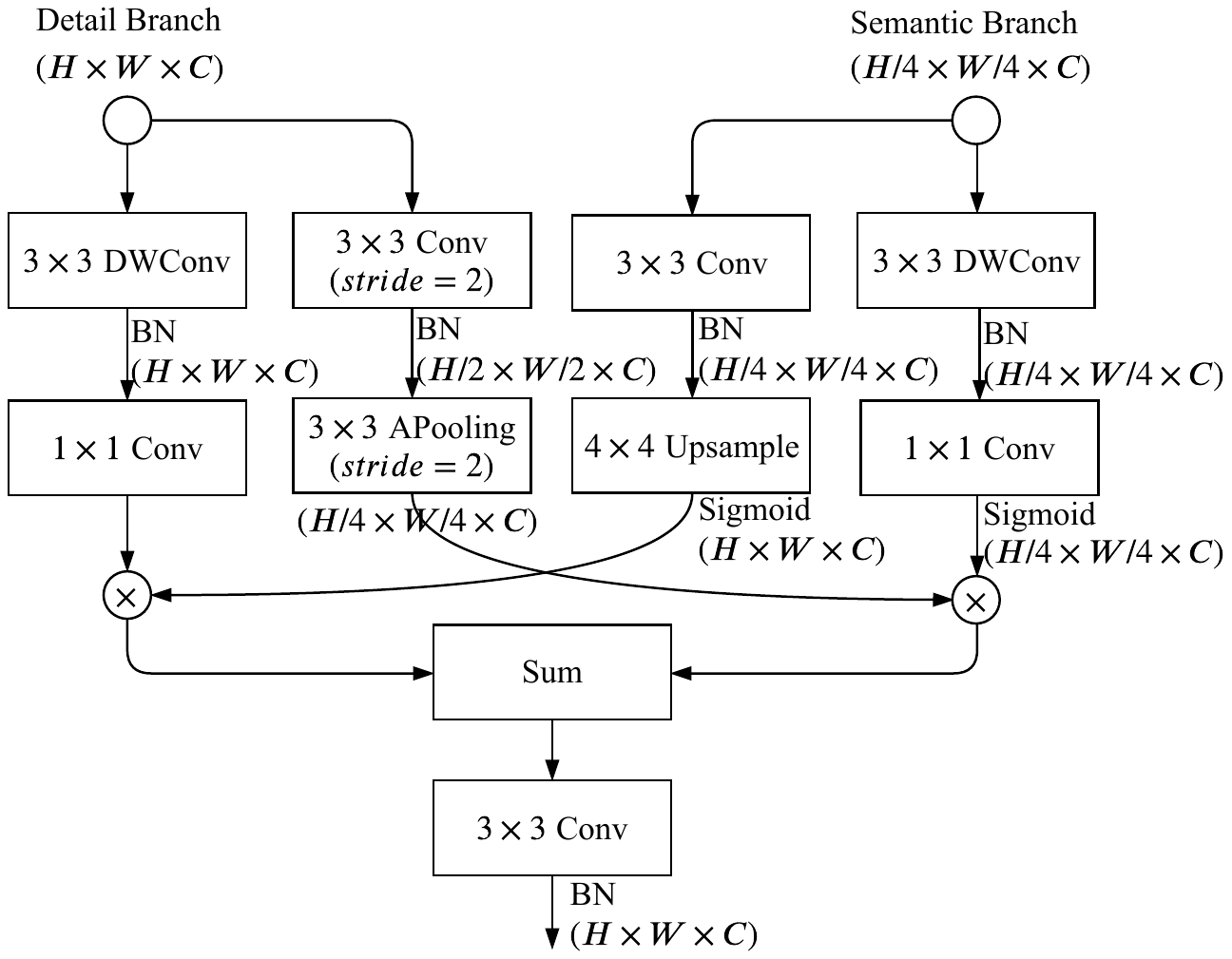}
\end{center}
\caption{\textbf{Detailed design of Bilateral Guided Aggregation Layer}.
Notation: \textit{Conv} is convolutional operation. \textit{DWConv} is depth-wise convolution. \textit{APooling} is average pooling. \textit{BN} denotes the batch normalization. \textit{Upsample} means bilinear interpolation. \textit{Sigmoid} is the Sigmoid activation function. \textit{Sum} means summation.
Meanwhile, $1\times1, 3\times3$ denote the kernel size, $H\times W\times C$ means the tensor shape (height, width, depth), $\bigotimes$ represents element-wise product.
}
\label{fig:aggregation-layer}
\end{figure}

\subsection{Bilateral Guided Aggregation}
\label{sec:instantiation:aggregation-layer}
There are some different manners to merge two types of feature response, \ie element-wise summation and concatenation.
However, the outputs of both branches have different levels of feature representation.
The Detail Branch is for the low-level, while the Semantic Branch is for the high-level.
Therefore, simple combination ignores the diversity of both types of information, leading to worse performance and hard optimization.

Based on the observation, we propose the Bilateral Guided Aggregation Layer to fuse the complementary information from both branches, as illustrated in Figure~\ref{fig:aggregation-layer}.
This layer employs the contextual information of Semantic Branch to guide the feature response of Detail Branch.
With different scale guidance, we can capture different scale feature representation, which inherently encodes the multi-scale information.
Meanwhile, this guidance manner enables efficient communication between both branches compared to the simple combination.

\begin{figure}[t]
\footnotesize
\centering
\renewcommand{\tabcolsep}{1pt} %
\ra{1} %
\begin{center}
\includegraphics[width=0.95\linewidth]{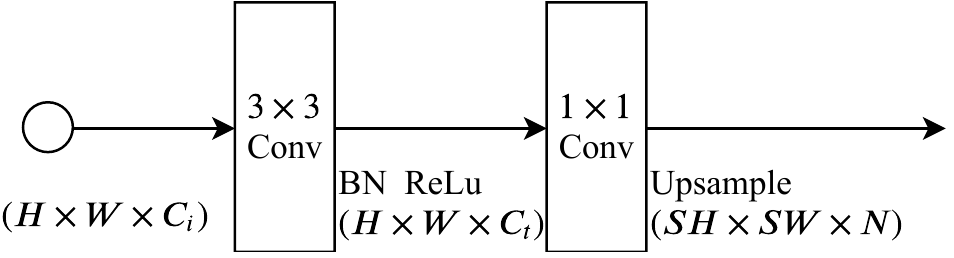}
\end{center}
\caption{\textbf{Detailed design of Segmentation Head in Booster}.
Notation: \textit{Conv} is convolutional operation. \textit{BN} denotes the batch normalization. \textit{Upsample} means bilinear interpolation.
Meanwhile, $1\times1, 3\times3$ denote the kernel size, $H\times W\times C$ means the tensor shape (height, width, depth), $C$ represents the channel dimension, $S$ denotes the scale ratio of upsampling, and $N$ is the final output dimension.
}
\label{fig:seg-head}
\end{figure}

\subsection{Booster Training Strategy}
\label{sec:instantiation:booster}
To further improve the segmentation accuracy, we propose a booster training strategy.
As the name implies, it is similar to the rocket booster: it can enhance the feature representation in the training phase and can be discarded in the inference phase.
Therefore, it increases little computation complexity in the inference phase.
As illustrated in Figure~\ref{fig:network}, we can insert the auxiliary segmentation head to different positions of the Semantic Branch. 
In Section~\ref{sec:ablation-city}, we analyze the effect of different positions to insert.
Figure~\ref{fig:seg-head} illustrates the details of the segmentation head.
We can adjust the computational complexity of auxiliary segmentation head and main segmentation head by controlling the channel dimension $C_t$. 

\section{Experimental Results}
\label{sec:experiment}
In this section, we first introduce the datasets and the implementation details. 
Next, we investigate the effects of each component of our proposed approach on City\-scap\-es validation set. 
Finally, we report our final accuracy and speed results on different benchmarks compared with other algorithms.

\paragraph{Datasets.}
Cityscapes~\citep{Cityscapes} focuses on semantic understanding of urban street scenes from a car perspective.
The dataset is split into training, validation and test sets, with $2,975$, $500$ and $1,525$ images respectively.
In our experiments, we only use the fine annotated images to validate the effectiveness of our proposed method.
The annotation includes 30 classes, 19 of which are used for semantic segmentation task.
This dataset is challenging for the real-time semantic segmentation because of its high resolution of $2,048\times1,024$.

Cambridge-driving Labeled Video Database (Cam\-Vid) \citep{CamVid} is a road scene dataset from the perspective of a driving automobile. 
It contains 701 images with $960\times720$ resolution extracted from the video sequences.
Following the pioneering work \citep{Brostow-ECCV-MotionPointClouds-2008, Sturgess-motionfeature-BMVC-2009,Badrinarayanan-PAMI-SegNet-2017}, the images are split into 367 for training, 101 for validation and 233 for testing. 
We use the subset of 11 classes of the provided 32 candidate categories for the fair comparison with other methods. 
The pixels do not belong to one of these classes are ignored.

COCO-Stuff~\citep{COCO-Stuff} augments 10K complex images of the popular COCO~\citep{Lin-COCO-2014} dataset with dense stuff annotations.
This is also a challenging dataset for the real-time semantic segmentation because it has more complex categories, including 91 thing and 91 stuff classes for evaluation.
For a fair comparison, we follow the split in \citep{COCO-Stuff}: 9K images for training and 1K images for testing.

\begin{table*}\centering\vspace{-1em}
\small
\tablestyle{4pt}{1.05}
\caption{\textbf{Ablations on Cityscapes}.
We validate the effectiveness of each component step by step.
We show segmentation accuracy (mIoU\%), and computational complexity measured in GFLOPs with the input of spatial size $2048\times1024$.
Notation: \textit{Detail} is the Detail Branch. \textit{Semantic} is the Semantic Branch. \textit{BGA} represents the Bilateral Guided Aggregation Layer. \textit{Booster} means the booster training strategy.
	\textit{OHEM} is the online hard example mining.
}
\label{tab:ablation}
\begin{tabular}{cc|ccc|c|c|c|c}
\shline
\multirow{2}{*}{Detail}            & \multirow{2}{*}{Semantic}             & \multicolumn{3}{c|}{Aggregation}                                              & \multirow{2}{*}{Booster} & \multirow{2}{*}{OHEM}            & \multirow{2}{*}{mIoU(\%)} & \multirow{2}{*}{GFLOPs} \\ 
\cline{3-5} &           & Sum        & Concate   & BGA &  &  &  & \\ \hline
\cmark & 			&            &  &  &  							& & 62.35 & 15.26 \\
           & \cmark &            &  &  &  							& & 64.68 & 7.63 \\ \hline
\cmark & \cmark & \cmark &  &  &  							& & 68.60 & 20.77 \\
\cmark & \cmark &            & \cmark &    &  				& & 68.93 & 21.98 \\
\cmark & \cmark &  &  				     & \cmark &  		 & & 69.67 & 21.15 \\ \hline
\cmark & \cmark &  &  					 & \cmark & \cmark & & 73.19 & 21.15 \\
\cmark & \cmark &  &  					 & \cmark & \cmark & \cmark & \textbf{73.36} & 21.15 \\
\shline
\end{tabular}
\end{table*}

\begin{table*}[t]\centering\vspace{-1em}
\captionsetup[subfloat]{captionskip=2pt}
\captionsetup[subffloat]{justification=centering}
\small
\caption{\textbf{Ablations on the Semantic Branch design on Cityscapes}.
We conduct experiments about the channel capacity, the block design, and the expansion ratio of the Semantic Branch.
Notation: \textit{GLayer} indicates the Gather Layer, the first $3\times3$ convolution in GE Layer. \textit{DDWConv} is double depth-wise convolution layer.
}
\label{tab:analysis}
\subfloat[\textbf{Channel capacity ratio}: Varying values of $\lambda$ can control the channel capacity of the first two stages in the Semantic Branch.
The channel dimensions of the last two stages are still $64$ and $128$.
Here, we choose $\lambda=1/4$.
\label{tab:analysis:channel-capacity}]{
\tablestyle{2.5pt}{1.05}
\begin{tabular}{r|c|c}
\shline
             & mIoU(\%)              		  & GFLOPs               \\ \hline
Detail-only  & 62.35                 		  & 15.26 \\ \hline
$\lambda=1/2$ & 69.66                 		  & 25.84 \\
$1/4$        & \textbf{69.67}                 & 21.15 \\
$1/8$        & 69.26                 		  & 19.93 \\
$1/16$       & 68.27			     		  & 19.61 \\ 
\shline
\multicolumn{3}{c}{} \\ %
\end{tabular}}\hspace{3mm}
\subfloat[\textbf{Block Analysis}: We specifically design the GE Layer and adopt double depth-wise convolutions when $stride=2$.
The second row means we use one $5\times5$ depth-wise convolution instead of two $3\times3$ depth-wise convolution.
The third row represents we replace the first $3\times3$ convolution layer of GE Layer with the $1\times1$ convolution.
\label{tab:analysis:block}]{
\tablestyle{2.5pt}{1.05}
\begin{tabular}{ccc|c|c}
\shline
GLayer    & DDWConv & Context & mIoU(\%) & GFLOPs \\ \hline
\cmark & \cmark & \cmark & \textbf{69.67} & 21.15 \\ \hline
\cmark & \cmark &            & 69.01          & 21.07 \\
\cmark &            & \cmark & 68.98          & 21.15 \\
           & \cmark & \cmark & 66.62          & 15.78 \\
\shline
\multicolumn{5}{c}{} \\ %
\multicolumn{5}{c}{} \\ %
\end{tabular}}\hspace{3mm}
\subfloat[\textbf{Expansion ratio}: Varying values of $\epsilon$ can affect the representative ability of the Semantic Branch.
We choose the $\epsilon=6$ to make the trade-off between accuracy and computation complexity.
\label{tab:analysis:expansion-ratio}]{
\tablestyle{2.5pt}{1.05}
\begin{tabular}{r|c|c}
\shline
             & mIoU(\%)              			 & GFLOPs               \\ \hline
Detail-only  & 62.35                 		     & 15.26  \\ \hline           
$\epsilon=1$  & 67.48                  			 & 17.78  \\
$2$          & 68.41                  			 & 18.45  \\
$4$          & 68.78                  			 & 19.8   \\
$6$          & \textbf{69.67}					 & 21.15  \\ 
$8$          & 68.99 				 			 & 22.49  \\ 
\shline
\end{tabular}}
\end{table*}

\paragraph{Training.}
Our models are trained from scratch with the ``kaiming normal'' initialization manner~\citep{He-ICCV-Rectifier-2015}.
We use the stochastic gradient descent (SGD) algorithm with 0.9 momentum to train our model. 
For all datasets, we adopt 16 batch size.
For the Cityscapes and CamVid datasets, the weight decay is 0.0005 weight decay while the weight decay is 0.0001 for the COCO-Stuff dataset.
We note that the weight decay regularization is only employed on the parameters of the convolution layers.
The initial rate is set to $5e^{-2}$ with a ``poly'' learning rate strategy in which the initial rate is multiplied by $(1 - \frac{iter}{iters_{max}})^{power}$ each iteration with power 0.9. 
Besides, we train the model for 150K, 10K, 20K iterations for the Cityscapes dataset, CamVid dataset, and COCO-Stuff datasets respectively.

For the augmentation, we randomly horizontally flip, randomly scale, and randomly crop the input images to a fixed size for training. 
The random scales contain \{ 0.75, 1, 1.25, 1.5, 1.75, 2.0\}. 
And the cropped resolutions are $2048\times1024$ for Cityscapes, $960\times720$ for CamVid, $640\times640$ for COCO-Stuff respectively.
Besides, the augmented input of Cityscapes will be resized into $1024\times512$ resolution to train our model.

\paragraph{Inference.}
We do not adopt any evaluation tricks, \eg sliding-window evaluation and multi-scale testing, which can improve accuracy but are time-consuming.
With the input of $2048\times1024$ resolution, we first resize it to $1024\times512$ resolution to inference and then resize the prediction to the original size of the input.
We measure the inference time with only one GPU card and repeat 5000 iterations to eliminate the error fluctuation.
We note that the time of resizing is included in the inference time measurement.
In other words, when measuring the inference time, the practical input size is $2048\times1024$.
Meanwhile, we adopt the standard metric of the mean intersection of union (mIoU) for the Cityscapes dataset and CamVid dataset, while the mIoU and pixel accuracy (pixAcc) for the COCO-Stuff dataset.

\paragraph{\rm \textbf{Setup.}}
We conduct experiments based on PyTorch 1.0.
The measurement of inference time is executed on one NVIDIA GeForce GTX 1080Ti with the CUDA 9.0, CUDNN 7.0 and {TensorRT v5.1.5}\footnote{We use FP32 data precision.}.

\subsection{Ablative Evaluation on Cityscapes}
\label{sec:ablation-city}
This section introduces the ablation experiments to validate the effectiveness of each component in our method.
In the following experiments, we train our models on Cityscapes \citep{Cityscapes} training set and evaluate on the Cityscapes validation set.

\begin{figure*}[t]
\footnotesize
\centering
\renewcommand{\tabcolsep}{1pt} %
\renewcommand{\arraystretch}{1} %
\begin{center}
\begin{tabular}{ccc}
\includegraphics[width=0.3\linewidth]{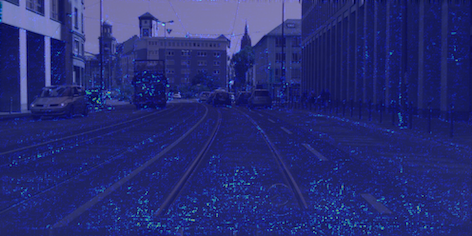} &
\includegraphics[width=0.3\linewidth]{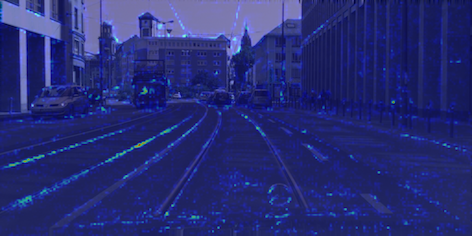} &
\includegraphics[width=0.3\linewidth]{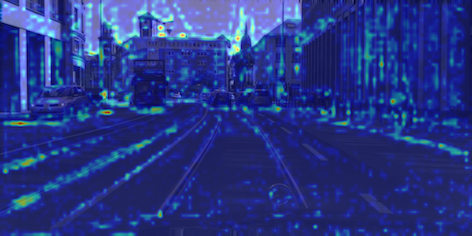} \\
\includegraphics[width=0.3\linewidth]{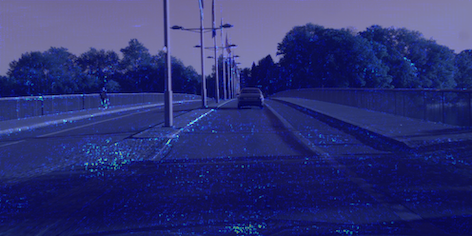} &
\includegraphics[width=0.3\linewidth]{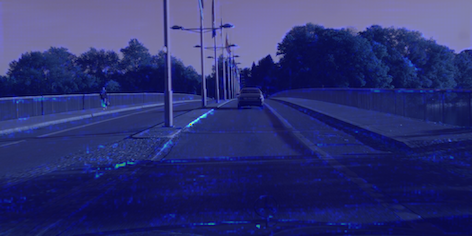} &
\includegraphics[width=0.3\linewidth]{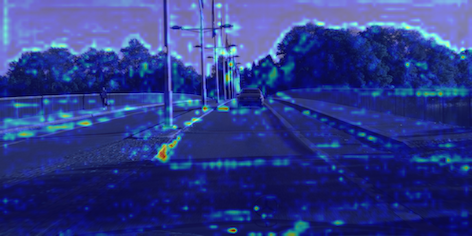} \\
\includegraphics[width=0.3\linewidth]{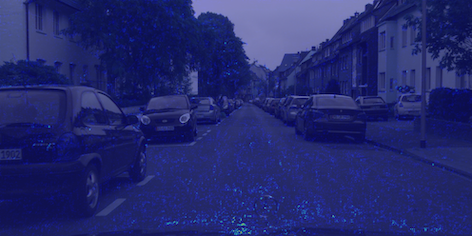} &
\includegraphics[width=0.3\linewidth]{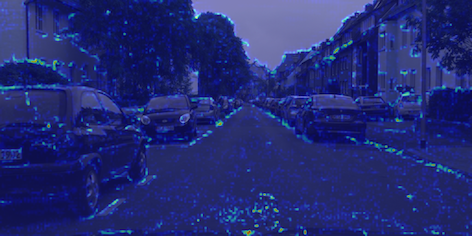} &
\includegraphics[width=0.3\linewidth]{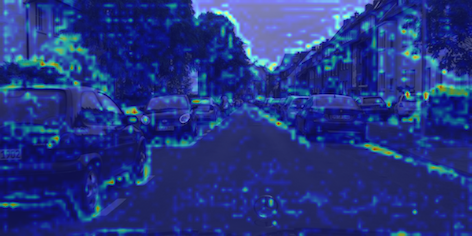} \\
\includegraphics[width=0.3\linewidth]{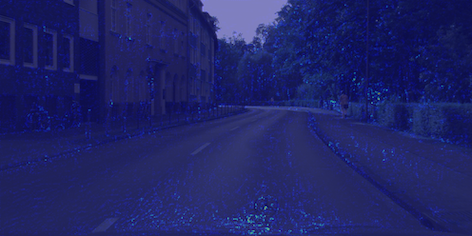} &
\includegraphics[width=0.3\linewidth]{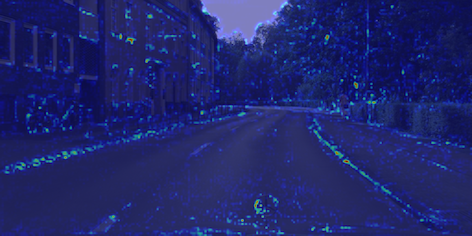} &
\includegraphics[width=0.3\linewidth]{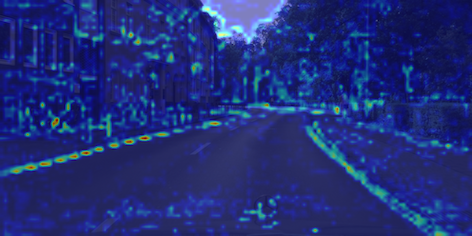} \\
(a) Stage$_1$ &
(b) Stage$_2$ &
(c) Stage$_3$ \\
\end{tabular}
\end{center}
\caption{\textbf{Examples showing visual explanations for the different stages of the Detail Branch.} Following the Grad-CAM~\citep{Selvaraju-ICCV-Grad-cam-2017}, we visualize the Grad-CAMs of Detail Branch. The visualization shows that Detail Branch can focus on the spatial details, \eg boundary, gradually.}
\label{fig:vis:details-change}
\end{figure*}

\paragraph{Individual pathways.} We first explore the effect of individual pathways specifically.
The first two rows in Table~\ref{tab:ablation} illustrates the segmentation accuracy and computational complexity of using only one pathway alone.
The Detail Branch lacks sufficient high-level semantics, while the Semantic Branch suffers from a lack of low-level spatial details, which leads to unsatisfactory results.
Figure~\ref{fig:vis:details-change} illustrates the gradual attention on the spatial details of Detail Branch.
The second group in Table~\ref{tab:ablation} shows that the different combinations of both branches are all better than the only one pathway models.
Both branches can provide a complementary representation to achieve better segmentation performance.
The Semantic Branch and Detail Branch alone only achieve $64.68\%$ and $62.35\%$ mean IoU.
However, with the simple summation, the Semantic Branch can bring in over $6\%$ improvement to the  Detail Branch, while the Detail Branch can acquire $4\%$ gain for the Semantic Branch.
This observation shows that the representations of both branches are complementary.

\begin{figure*}[t]
\footnotesize
\centering
\renewcommand{\tabcolsep}{1pt} %
\renewcommand{\arraystretch}{1} %
\begin{center}
\begin{tabular}{ccccc}
\includegraphics[width=0.2\linewidth]{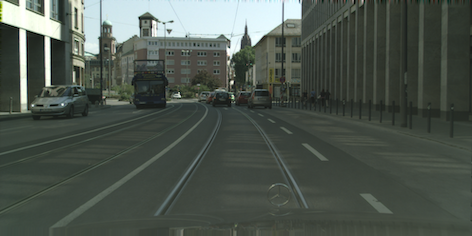} &
\includegraphics[width=0.2\linewidth]{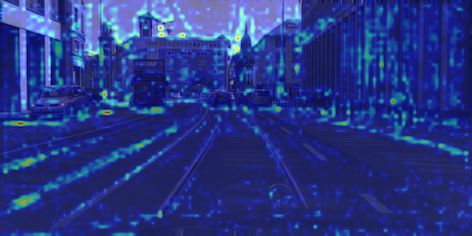} &
\includegraphics[width=0.2\linewidth]{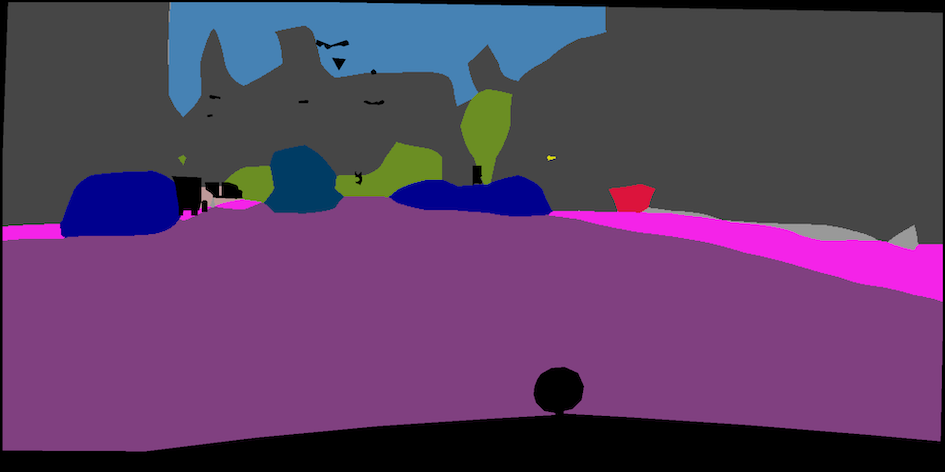} &
\includegraphics[width=0.2\linewidth]{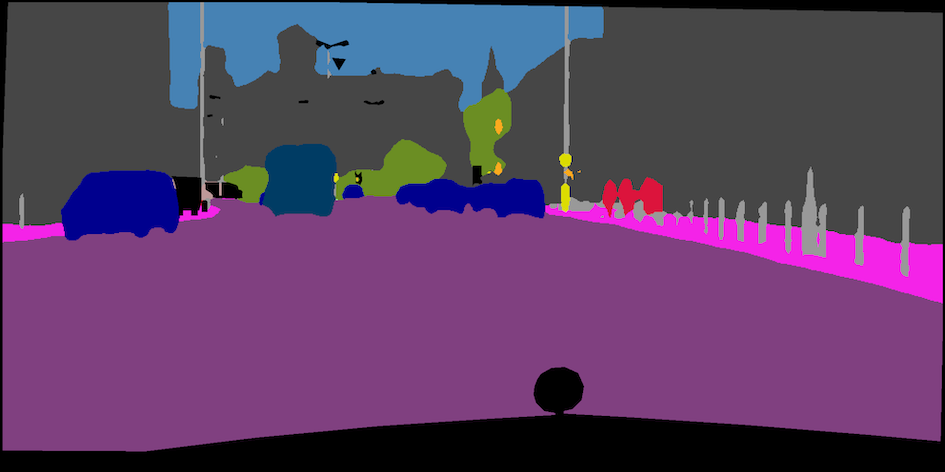} &
\includegraphics[width=0.2\linewidth]{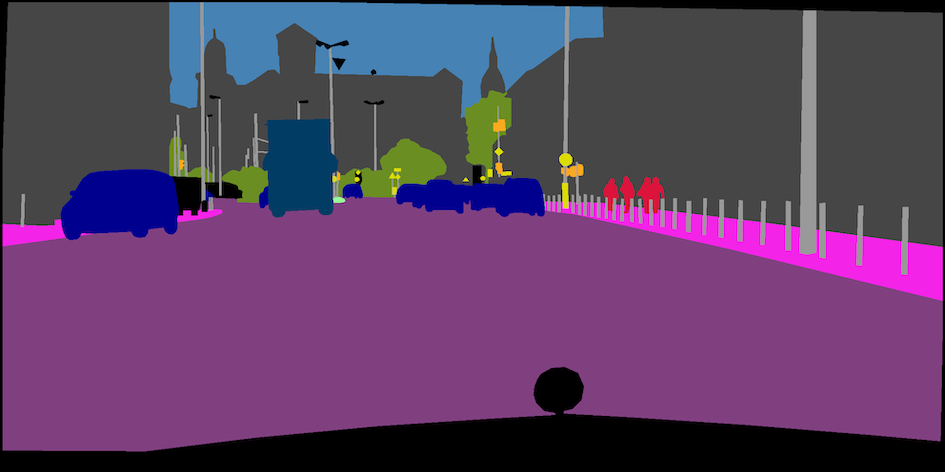} \\
\includegraphics[width=0.2\linewidth]{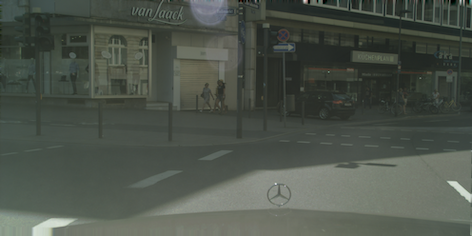} &
\includegraphics[width=0.2\linewidth]{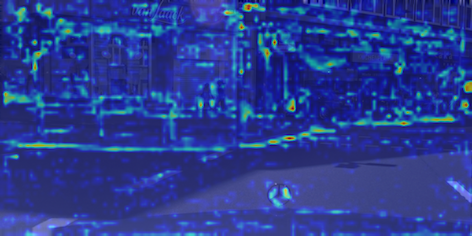} &
\includegraphics[width=0.2\linewidth]{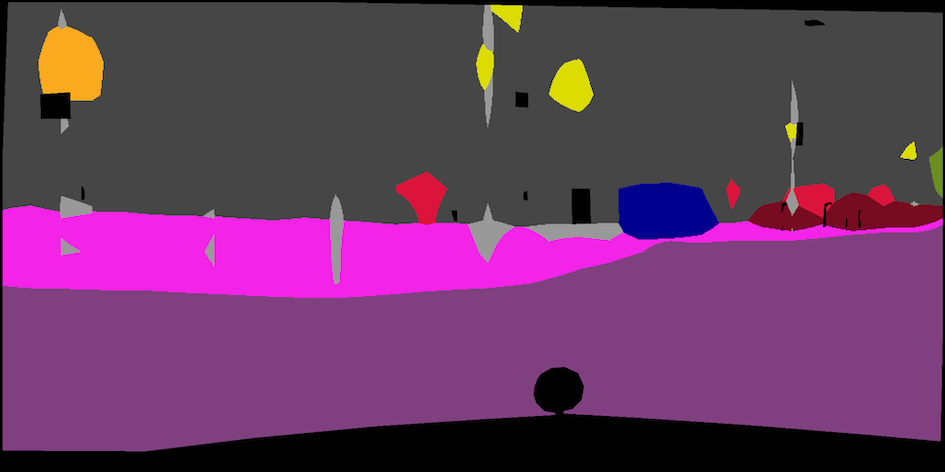} &
\includegraphics[width=0.2\linewidth]{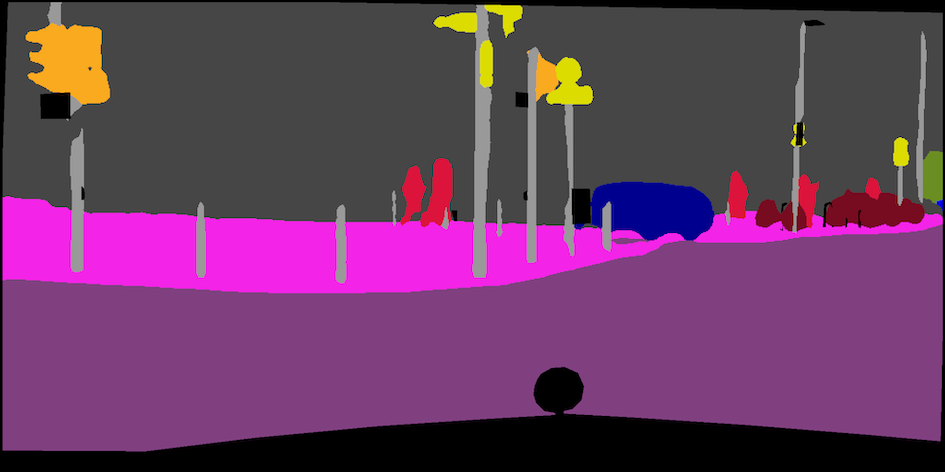} &
\includegraphics[width=0.2\linewidth]{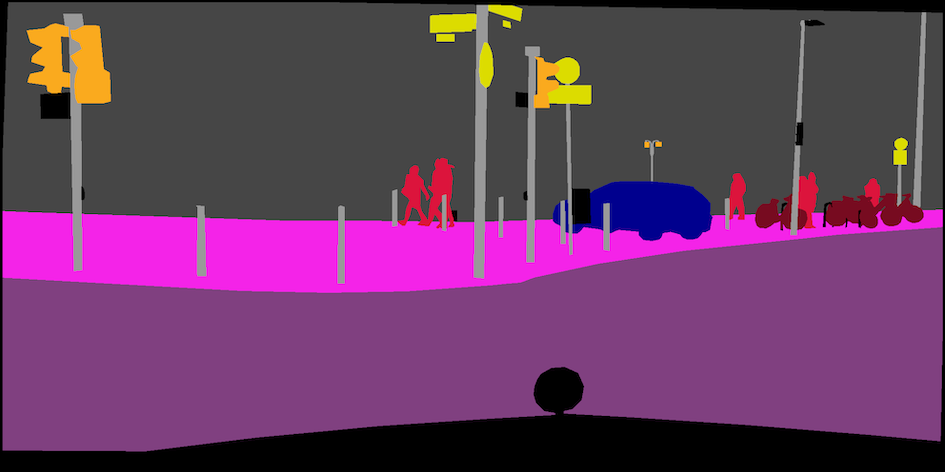} \\
\includegraphics[width=0.2\linewidth]{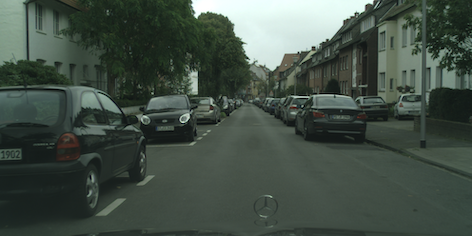} &
\includegraphics[width=0.2\linewidth]{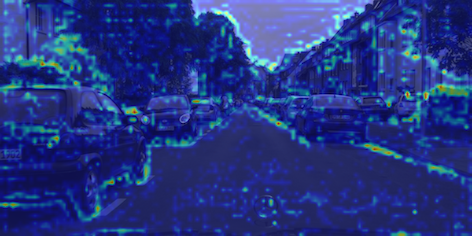} &
\includegraphics[width=0.2\linewidth]{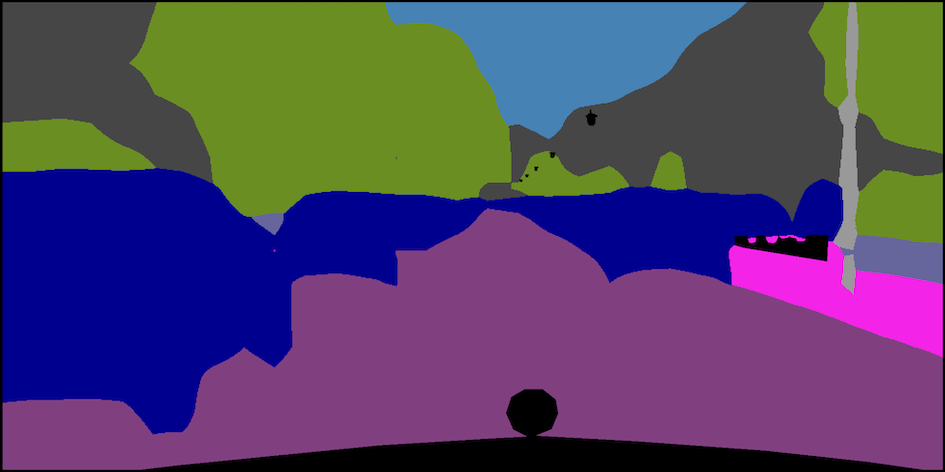} &
\includegraphics[width=0.2\linewidth]{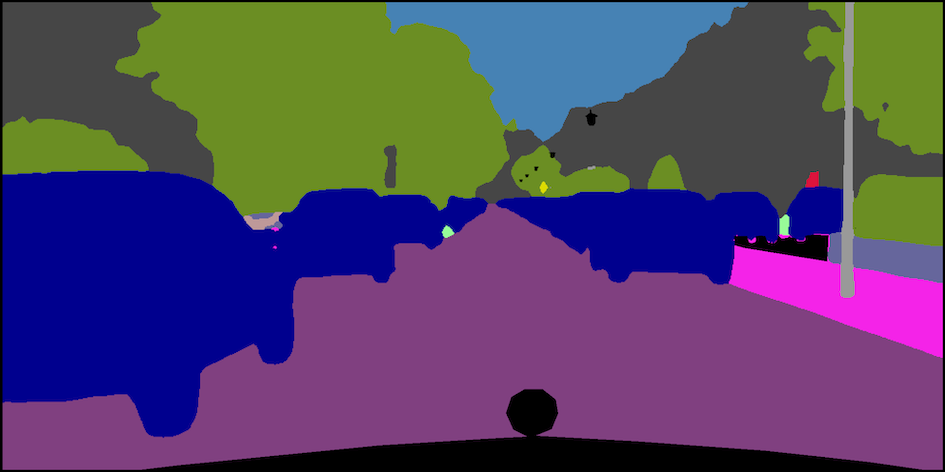} &
\includegraphics[width=0.2\linewidth]{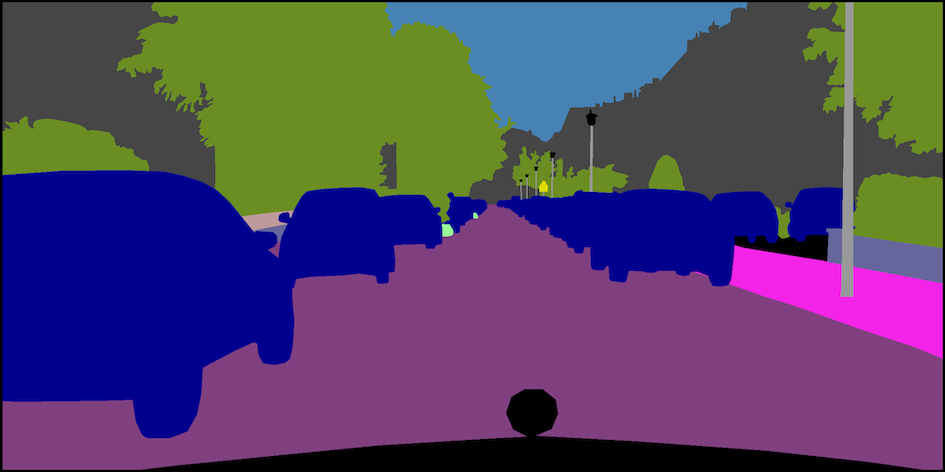} \\
\includegraphics[width=0.2\linewidth]{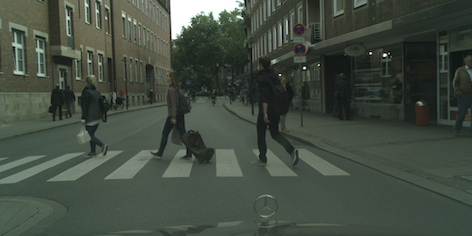} &
\includegraphics[width=0.2\linewidth]{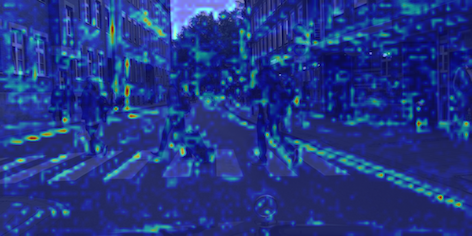} &
\includegraphics[width=0.2\linewidth]{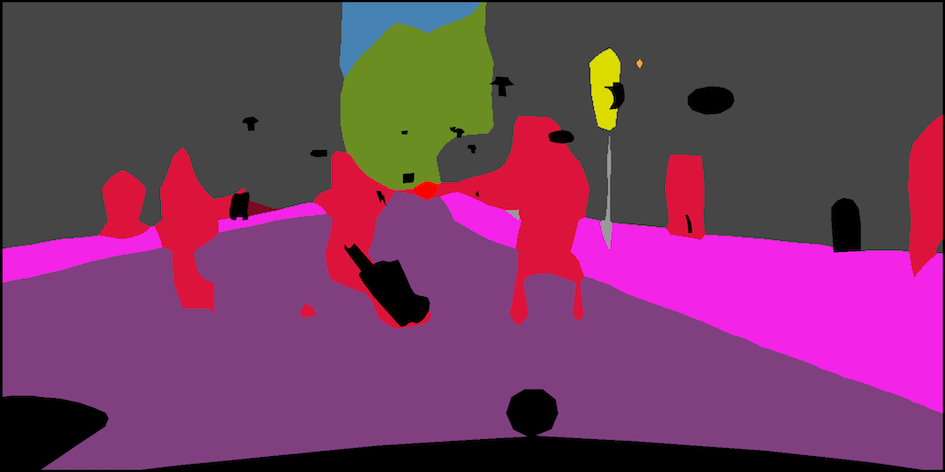} &
\includegraphics[width=0.2\linewidth]{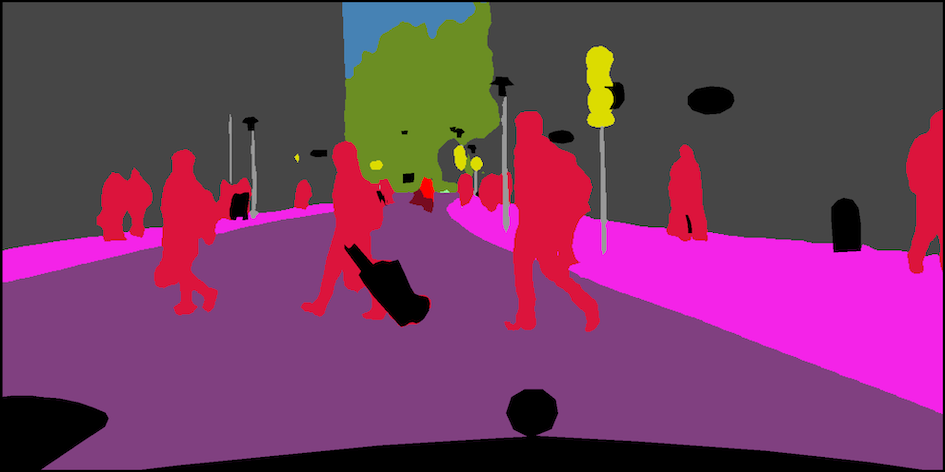} &
\includegraphics[width=0.2\linewidth]{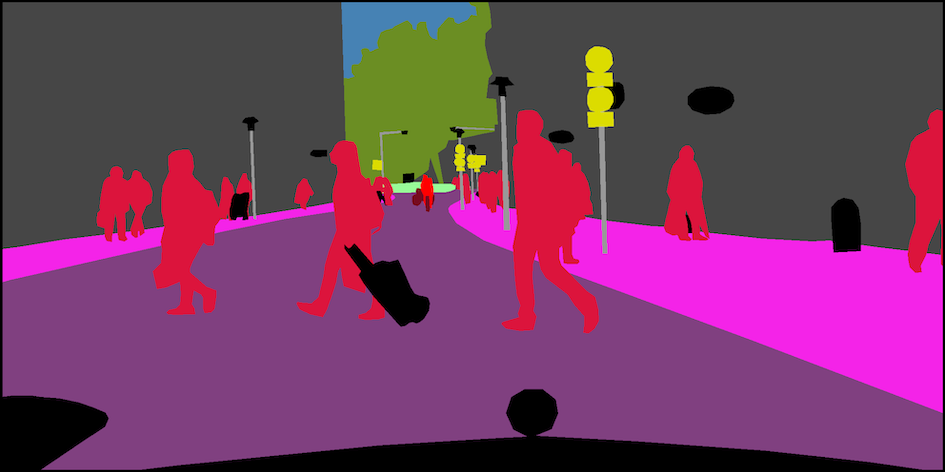} \\
(a) Input &
(b) Detail Branch &
(c) Semantic Branch &
(d) Prediction &
(e) Groundtruth \\
\end{tabular}
\end{center}
\caption{\textbf{Visual improvement of the Bilateral Guided Aggregation layer on the Cityscapes \emph{val} set.}}
\label{fig:vis:aggregation}
\end{figure*}

\paragraph{Aggregation methods.} We also investigate the aggregation methods of two branches, as illustrated in Table~\ref{tab:ablation}.
For an effective and efficient aggregation, we design the Bilateral Guided Aggregation Layer, which adopts the high-level semantics as the guidance to aggregate the multi-scale low-level details.
We also show two variants without Bilateral Guided Aggregation Layer as the naive aggregation baseline: \textit{summation} and \textit{concatenation} of the outputs of both branches.
For a fair comparison, the inputs of the \textit{summation} and \textit{concatenation} are through one separable layer respectively.
Figure~\ref{fig:vis:aggregation} demonstrates the visualization outputs of Detail Branch, Semantic Branch and the aggregation of both branches.
This illustrates that Detail Branch can provide sufficient spatial details, while Semantic Branch captures the semantic context.

Table~\ref{tab:analysis} illustrates a series of analysis experiments on the Semantic Branch design.

\paragraph{Channel capacity of Semantic Branch.} As discussed in Section~\ref{sec:concept:semantic-branch} and Section~\ref{sec:instantiation:semantic-branch}, the Semantic Branch is responsible for the high-level semantics, without caring the spatial details.
Therefore, the Semantic Branch can be made very lightweight with low channel capacity, which is adapted by the channel capacity ratio of $\lambda$. 
Table~\ref{tab:analysis:channel-capacity} illustrates the detailed comparison experiments of varying $\lambda$.

Different $\lambda$ brings in different extents of improvement to the Detail-only baseline.
Even when $\lambda = 1/16$, the first layer of the Semantic Branch has only 4 channel dimension, which also brings in $6\%$ ($62.35\%$ $\rightarrow$ $68.27\%$) improvement to the baseline.
Here, we employ $\lambda = 1/4$ as our default.

\begin{table}\centering\vspace{-1em}
\small
\tablestyle{3pt}{1.05}
\caption{\textbf{Booster position}.
We can add the auxiliary segmentation head to different positions as the booster of the Semantic Branch. 
Here, stage$_s$ represents the auxiliary segmentation head can be added after $s$ stage.
stage$_{5\_4}$ and stage$_{5\_5}$ means the position before or after the Context Embedding Block respectively.
OHEM represents the online bootstrapping strategy.
}
\label{tab:booster}
\begin{tabular}{ccccc|c|c}
\shline
stage$_2$ & stage$_3$ & stage$_4$ & stage$_{5\_4}$ & stage$_{5\_5}$ & OHEM & mIoU(\%) \\ \hline
   		   &    		&    		 &    		  &            &  & 69.67 \\ \hline
\cmark & \cmark & \cmark & \cmark & \cmark &  & 73.04 \\
\cmark & \cmark & \cmark & \cmark & 		   &  & \underline{73.19} \\
\cmark & \cmark & \cmark & 			  & 		   &  & 71.62 \\
           & \cmark & \cmark & \cmark & \cmark &  & 72.84 \\
           &            & \cmark & \cmark & \cmark &  & 72.68 \\
 		   &  			&  			 & \cmark & \cmark &  & 72.03 \\
 \hline
 \cmark & \cmark & \cmark & \cmark & 		   & \cmark & \textbf{73.36} \\
 \shline

\end{tabular}
\end{table}

\paragraph{Block design of of Semantic Branch.} Following the pioneer work~\citep{Sandler-CVPR-MobileNetv2-2018, Howard-ICCV-Mobilenetv3-2019}, we design a Gather-and-Expansion Layer, as discussed in Section~\ref{sec:instantiation:semantic-branch} and illustrated in Figure~\ref{fig:gelayer}.
The main improvements consist of two-fold: (\romannumeral1) we adopt one $3\times3$ convolution as the Gather Layer instead of one point-wise convolution in the inverted bottleneck of MobileNetV2~\citep{Sandler-CVPR-MobileNetv2-2018}; (\romannumeral2) when $stride=2$, we employs two $3\times3$ depth-wise convolution to substitute a $5\times5$ depth-wise convolution.

Table~\ref{tab:analysis:block} shows the improvement of our block design.
The Gather-and-Expansion Layer can enlarge the receptive field to capture high-level semantics efficiently.

\begin{table}[t]\centering\vspace{-1em}
\captionsetup[subfloat]{captionskip=2pt}
\captionsetup[subffloat]{justification=centering}
\small
\tablestyle{2.5pt}{1.05}
\caption{\textbf{Generalization to large models}.
We enlarge our models from two aspects: (\romannumeral1) wider models; (\romannumeral2) deeper models.
}
\label{tab:large-model}
\subfloat[\textbf{Wider models}: Varying values of $\alpha$ can control the channel capacity of our architecture.
\label{tab:wider-model}]{
\begin{tabular}{r|c|c}
\shline
Wider & mIoU(\%)    & GFLOPs           \\ \hline
$\alpha=1.0$ & 73.36 & 21.15                 \\ \hline
$1.25$       & 73.61 & 34.98                 \\
$1.50$       & 74.67 & 49.46                 \\
$1.75$       & 74.04 & 66.45				 \\ 
$2.0$        & \textbf{75.11} & 85.94				 \\ \shline
\end{tabular}}\hspace{1mm}
\subfloat[\textbf{Deeper models}: Varying values of $d$ represents the layer number of the model.\label{tab:deeper-model}]{
\tablestyle{2pt}{1.05}
\begin{tabular}{r|c|c}
\shline
Deeper & mIoU(\%)    & GFLOPs \\ \hline       
$d=1.0$       & 73.36 & 21.15 \\ \hline
$2.0$         & 74.10 & 25.26 \\
$3.0$         & \textbf{74.28} & 29.38 \\
$4.0$         & 74.02 & 33.5  \\
\shline
\multicolumn{3}{c}{} \\ %
\end{tabular}}
\end{table}

\begin{table}\centering\vspace{-1em}
\small
\tablestyle{3pt}{1.05}
\caption{\textbf{Compatibility with other models}.
We empoly different light-weight models as the Semantic Branch to explore the compatibility of our architecture.
}
\label{tab:compatibility}
\begin{tabular}{l|c|c|c}
\shline
\multicolumn{1}{c|}{Semantic Branch} & Pretrained & mIoU(\%) & GFLOPs \\ \hline
ShuffleNetV2 $1.5\times$ & ImageNet & 74.07 & 128.71 \\
MobileNetV2  	         & ImageNet & 72.95  & 129.45 \\
ResNet-18 	             & ImageNet & 75.22  & 143.34 \\ \hline
Ours($\alpha=1.0,d=1.0$) & no   & 73.36 & 21.15 \\
Ours($\alpha=2.0,d=3.0$) & no   & \textbf{75.80} & 118.51 \\
\shline
\end{tabular}
\end{table}

\paragraph{Expansion ratio of GE layer.} The first $3\times3$ convolution layer in GE Layer is also an expansion layer, which can project the input to a high-dimensional space.
It has an advantage in memory access cost~\citep{Sandler-CVPR-MobileNetv2-2018, Howard-ICCV-Mobilenetv3-2019}.
The expansion ratio of $\epsilon$ can control the output dimension of this layer.
Table~\ref{tab:analysis:expansion-ratio} investigates the effect of varying $\epsilon$.
It is surprising to see that even with $\epsilon = 1$, the Semantic Branch can also improve the baseline by $4\%$ ($62.35\%$ $\rightarrow$ $67.48\%$) mean IoU, which validates the lightweight Semantic Branch is efficient and effective.

\paragraph{Booster training strategy.} We propose a booster training strategy to further improve segmentation accuracy, as discussed in Section~\ref{sec:instantiation:booster}.
We insert the segmentation head illustrated in Figure~\ref{fig:seg-head} to different positions of Semantic Branch in the training phase, which are discarded in the inference phase.
Therefore, they increase little computation complexity in the inference phase, which is similar to the booster of the rocket.
Table~\ref{tab:booster} shows the effect of different positions to insert segmentation head.
As we can see, the booster training strategy can obviously improve segmentation accuracy.
We choose the configuration of the third row of Table~\ref{tab:booster}, which further improves the mean IoU by over $3\%$ ($69.67\%$ $\rightarrow$ $73.19\%$), without sacrificing the inference speed.
Based on this configuration, we adopt the online bootstrapping strategy~\citep{Wu-Arxiv-HighPerf-2016} to improve the performance further.

\subsection{Generalization Capability}
\label{sec:large-models}
In this section, we mainly explore the generalization capability of our proposed architecture.
First, we investigate the performance of a wider model and deeper model in Table~\ref{tab:large-model}.
Next, we replace the Semantic Branch with some other general light-weight models to explore the compatibility ability in Table~\ref{tab:compatibility}.

\begin{table}[t]\centering
\small
\tablestyle{2pt}{1.2}
\caption{\textbf{Comparison with state-of-the-art on Cityscapes}. 
We train and evaluate our models with $2048\times1024$ resolution input, which is resized into $1024\times512$ in the model.
The inference time is measured on one NVIDIA GeForce 1080Ti card.
Notation: $\gamma$ is the downsampling ratio corresponding to the original $2048\times1024$ resolution. \textit{backbone} indicates the backbone models pre-trained on the ImageNet dataset.
``-'' represents that the methods do not report the corresponding result. The DFANet A and DFANet B adopt the $1024\times1024$ input size and use the optimized depth-wise convolutions to accelerate speed.}
\label{tab:city-performance-comp}
\begin{tabular}{l|c|c|c|cc|c}
\shline
\multicolumn{1}{c|}{\multirow{2}{*}{method}} & \multirow{2}{*}{ref.} &\multirow{2}{*}{$\gamma$} & \multirow{2}{*}{backbone} & \multicolumn{2}{c|}{mIoU(\%)} & \multirow{2}{*}{FPS} \\
\cline{5-6} & & & & \textit{val} & \textit{test}& \\
\hline
\multicolumn{6}{l}{\textit{large models}} \\ \hline
 CRF-RNN$^{*}$  & ICCV2015 &0.5  & VGG16 	  & - 	 & 62.5  & 1.4  \\
 DeepLab$^{*}$ 	& ICLR2015 &0.5  & VGG16    & -	 & 63.1  & 0.25 \\
 FCN-8S$^{*}$   & CVPR2015 &1.0  & VGG16    & -      & 65.3  & 2    \\
 Dilation10 	& ICLR2016 &1.0 & VGG16     & 68.7 & 67.1  & 0.25 \\
 LRR			& ECCV2016 &1.0 & VGG16     & 70.0 & 69.7  & -    \\
 Deeplabv2      & ICLR2016 &1.0 & ResNet101 & 71.4 & 70.4  & -    \\
 FRRN 			& CVPR2017 &0.5  & no 	  & -    & 71.8  & 2.1  \\
 RefineNet		& CVPR2017 &1.0 & ResNet101 & -    & 73.6  & -    \\
 DUC			& WACV2018 &1.0 & ResNet101 & 76.7 & \underline{76.1} & - \\
 PSPNet         & CVPR2017 &1.0 & ResNet101 & -    & \textbf{78.4}  & 0.78 \\
 \hline
 \multicolumn{6}{l}{\textit{small models}} \\ \hline
 ENet 	     & arXiv2016 &0.5   & no     	    & -    & 58.3  & 76.9  \\
 SQ          & NIPSW2016 & 1.0  & SqueezeNet 	& -    & 59.8  & 16.7  \\
 ESPNet      & ECCV2018 & 0.5   & ESPNet        & -    & 60.3  & 112.9 \\
 ESPNetV2    & CVPR2019 & 0.5   & ESPNetV2      & 66.4 & 66.2  &  -    \\
 ERFNet      & TITS2018 & 0.5   & no            & 70.0 & 68.0  & 41.7  \\
 Fast-SCNN   & BMVC2019 & 1.0   & no			& 68.6 & 68.0 & 123.5 \\
 ICNet       & ECCV2018 & 1.0	& PSPNet50		& -	   & 69.5 & 30.3 \\
 DABNet      & BMVC2019 & 1.0   & no		    & -    & 70.1  & 27.7 \\
 DFANet B    & CVPR2019 & 0.5$^*$ & Xception B    & -     & 67.1 & 120   \\
 DFANet A$'$ & CVPR2019 & 0.5   & Xception A    & -    & 70.3 & \textbf{160}   \\
  DFANet A & CVPR2019 & 0.5$^*$   & Xception A    & -    & 71.3 & 100   \\
 GUN       & BMVC2018 & 0.5   & DRN-D-22     	& 69.6 & 70.4 & 33.3  \\
 SwiftNet  & CVPR2019 & 1.0 & ResNet18		& 75.4 & 75.5 & 39.9 \\
 BiSeNetV1 & ECCV2018 & 0.75 & Xception39  & 69.0 & 68.4  & 105.8 \\
 BiSeNetV1 & ECCV2018 & 0.75 & ResNet18    & \underline{74.8} & 74.7  & 65.5 \\
 \hline
BiSeNetV2   & {---} & 0.5 & no       & 73.4 & 72.6 & \underline{156} \\
BiSeNetV2-L & {---} & 0.5 & no & \textbf{75.8} & 75.3 & 47.3  \\
\shline
\end{tabular}
\end{table}

\paragraph{Generalization to large models.}
Although our architecture is designed mainly for the light-weight task, \eg real-time semantic segmentation, BiSeNet V2 can be also generalized to large models.
We mainly enlarge the architecture from two aspects: (\romannumeral1) wider models, controlled by the width multiplier $\alpha$; (\romannumeral2) deeper models, controlled by the depth multiplier $d$.
Table~\ref{tab:large-model} shows the segmentation accuracy and computational complexity of wider models with the different width multiplier $\alpha$ and different depth multiplier $d$.
According to the experiments, we choose $\alpha=2.0$ and $d=3.0$ to build our large architecture, termed BiSeNetV2-Large, which achieves $75.8\%$ mIoU and GFLOPs.

\paragraph{Compatibility with other models.}
The BiSeNetV2 is a generic architecture with two branches.
In this work, we design some specific blocks for the Semantic Branch.
The Semantic Branch can be any light-weight convolutional models~\citep{He-CVPR-ResNet-2016, Howard-Arxiv-MobileNet-2017}.
Therefore, to explore the compatibility ability of our architecture, we conduct a series of experiments with different general light-weight models.
Table~\ref{tab:compatibility} shows the results of the combination with different models.

\begin{table}[t]
\centering
\small
\tablestyle{2pt}{1.2}
\caption{\textbf{Comparison with state-of-the-art on CamVid}.
With $960\times720$ input, we evaluate the segmentation accuracy and corresponding inference speed.
Notation: \textit{backbone} means the backbone models pre-trained on extra datasets, \eg ImageNet dataset and Cityscapes dataset.
$*$ indicates the models are pre-trained on Cityscapes.
$\dagger$ represents the models are trained from scratch.
}
\label{tab:camvid-performance-comp}
\begin{tabular}{l|c|c|c|c}
\shline
\multicolumn{1}{c|}{method} & ref. & backbone & mIoU(\%) & FPS \\
\hline
\multicolumn{4}{l}{\textit{large models}} \\ \hline
SegNet 		 & TPAMI2017 & VGG16      & 60.1  & 4.6  \\
DPN 		 & ICCV2015 & VGG16	  & 60.1  & 1.2  \\
Deeplab      & ICLR2015 & VGG16      & 61.6  & 4.9  \\
RTA	   		 & ECCV2018 & VGG16  	  & 62.5  & 0.2  \\
Dilation8 	 & ICLR2016 & VGG16  	  & 65.3  & 4.4  \\
PSPNet		 & CVPR2017 & ResNet50   & 69.1  & 5.4  \\
DenseDecoder & CVPR2018 & ResNeXt101 & 70.9  & -     \\
VideoGCRF$^*$& CVPR2018 & ResNet101  & 75.2  & -     \\
\hline
\multicolumn{4}{l}{\textit{small models}} \\ \hline
ENet 			& arXiv2016 & no		  & 51.3 & 61.2 \\
DFANet B        & CVPR2019 & Xception B  & 59.3 & \underline{160} \\
DFANet A        & CVPR2019 & Xception A  & 64.7 & 120 \\
ICNet			& ECCV2018 & PSPNet50	  & 67.1 & 27.8 \\
SwiftNet		& CVPR2019 & ResNet18$^\dagger$ & 63.33 & - \\
SwiftNet		& CVPR2019 & ResNet18	  & 72.58 & - \\
BiSeNetV1		& ECCV2018 & Xception 39 & 65.6 & \textbf{175} \\
BiSeNetV1		& ECCV2018 & ResNet18	  & 68.7 & 116.25 \\
\hline
BiSeNetV2       & {---} & no  & 72.4 & 124.5  \\
BiSeNetV2-L     & {---} & no  & 73.2 & 32.7  \\ 
BiSeNetV2$^*$   & {---} & no  & \underline{76.7} & 124.5 \\
BiSeNetV2-L$^*$ & {---} & no  & \textbf{78.5} & 32.7 \\
\shline
\end{tabular}
\end{table}

\subsection{Performance Evaluation}
In this section, we compare our best model (BiSeNetV2 and BiSeNetV2-Large) with other \emph{state-of-the-art} methods on three benchmark datasets: Cityscapes, CamVid and COCO-Stuff.

\begin{figure*}[t]
\footnotesize
\centering
\renewcommand{\tabcolsep}{1pt} %
\renewcommand{\arraystretch}{1} %
\begin{center}
\begin{tabular}{cccc}
\includegraphics[width=0.24\linewidth]{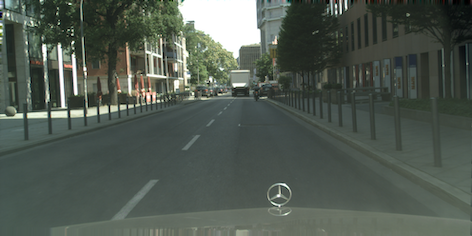} &
\includegraphics[width=0.24\linewidth]{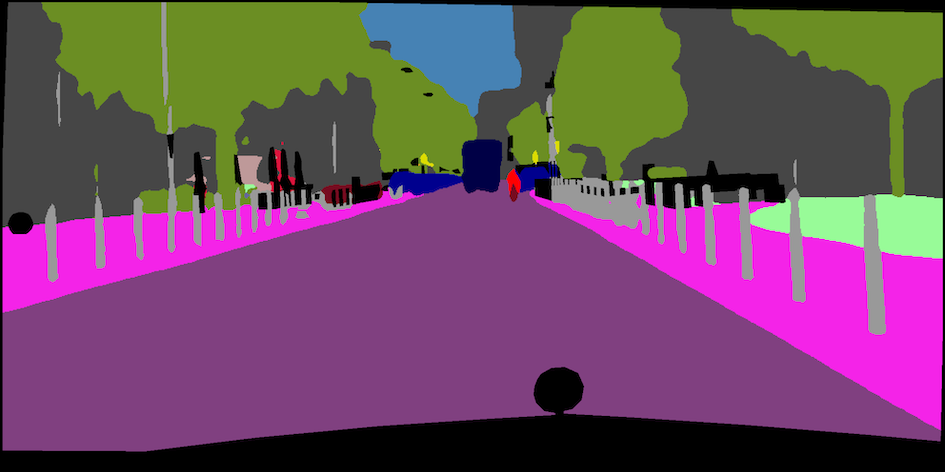} &
\includegraphics[width=0.24\linewidth]{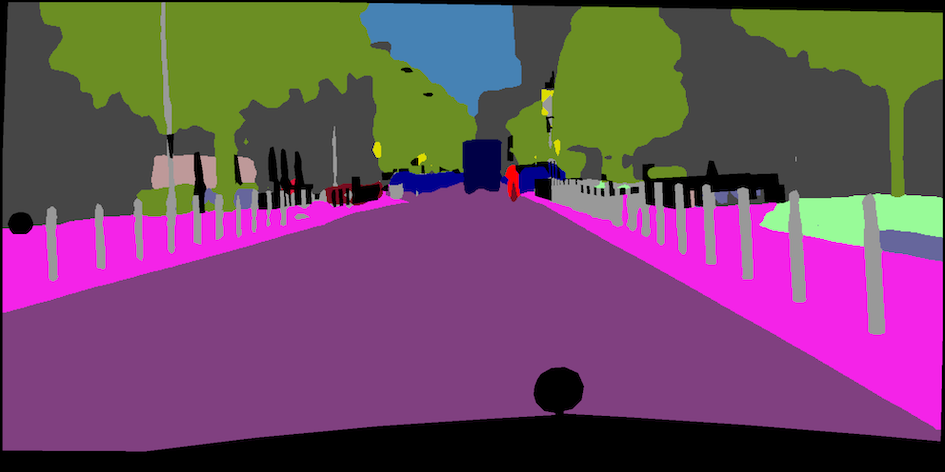} &
\includegraphics[width=0.24\linewidth]{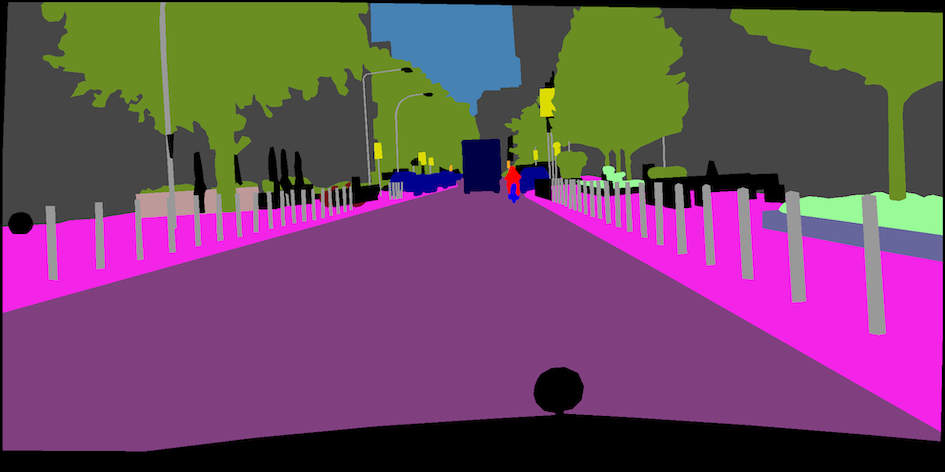} \\
\includegraphics[width=0.24\linewidth]{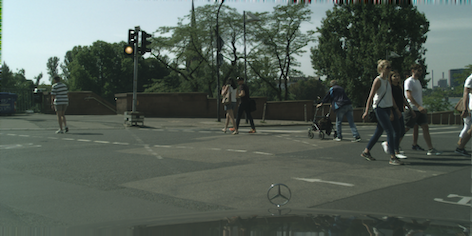} &
\includegraphics[width=0.24\linewidth]{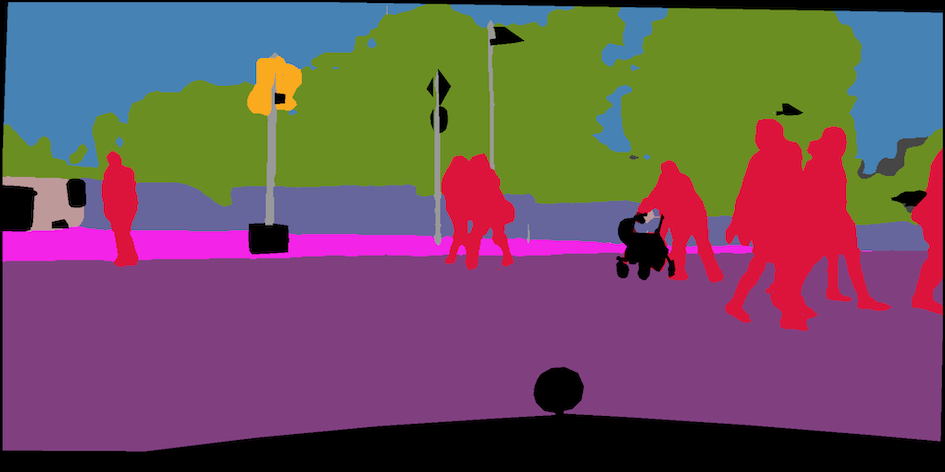} &
\includegraphics[width=0.24\linewidth]{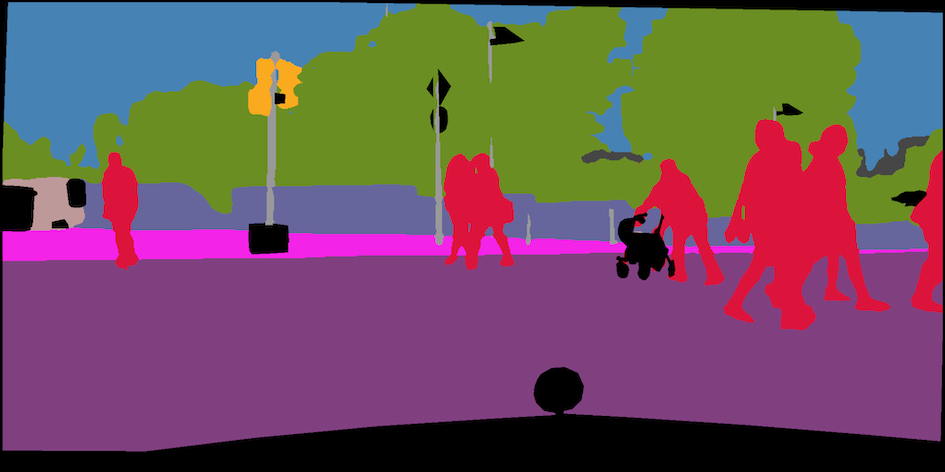} &
\includegraphics[width=0.24\linewidth]{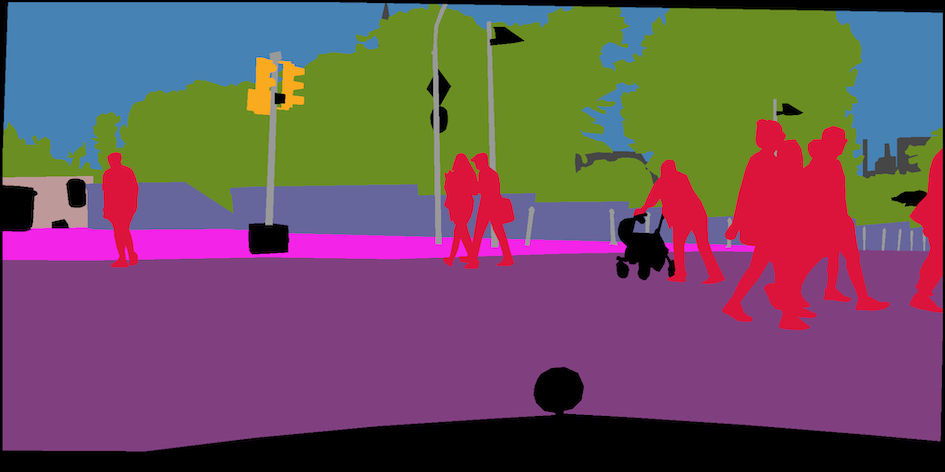} \\
\includegraphics[width=0.24\linewidth]{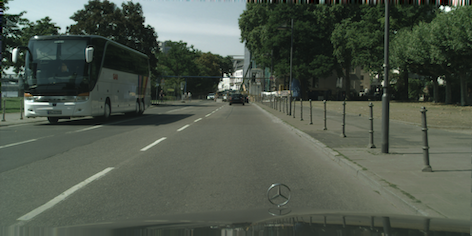} &
\includegraphics[width=0.24\linewidth]{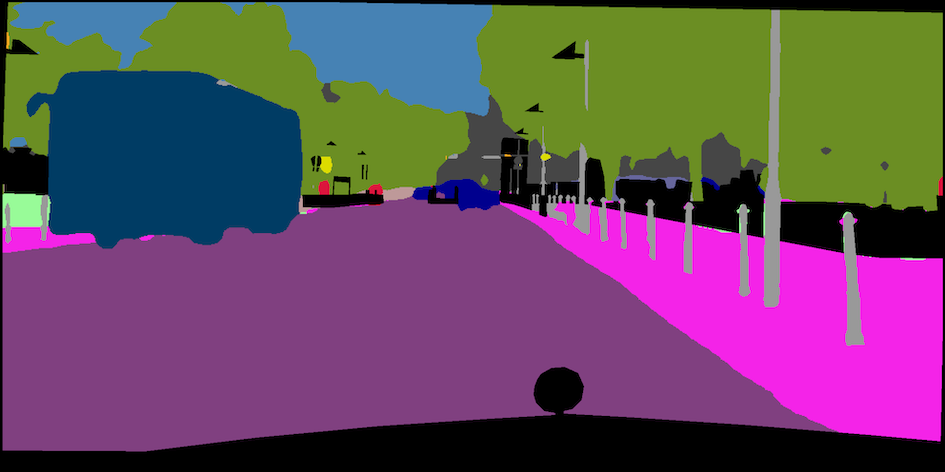} &
\includegraphics[width=0.24\linewidth]{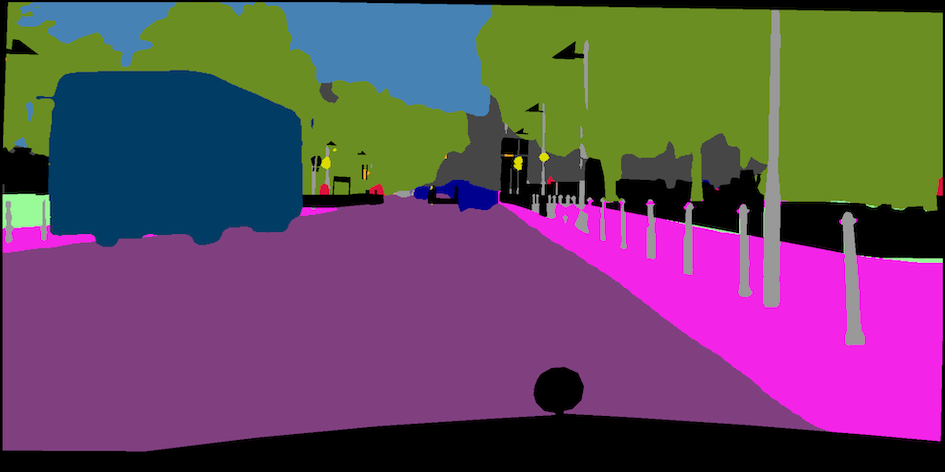} &
\includegraphics[width=0.24\linewidth]{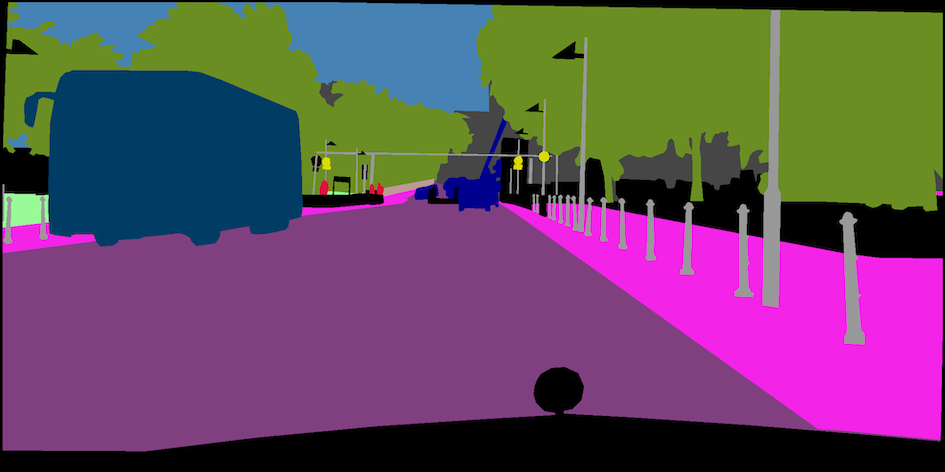} \\
\includegraphics[width=0.24\linewidth]{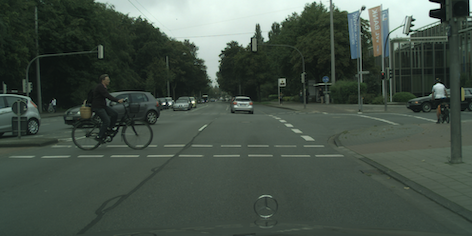} &
\includegraphics[width=0.24\linewidth]{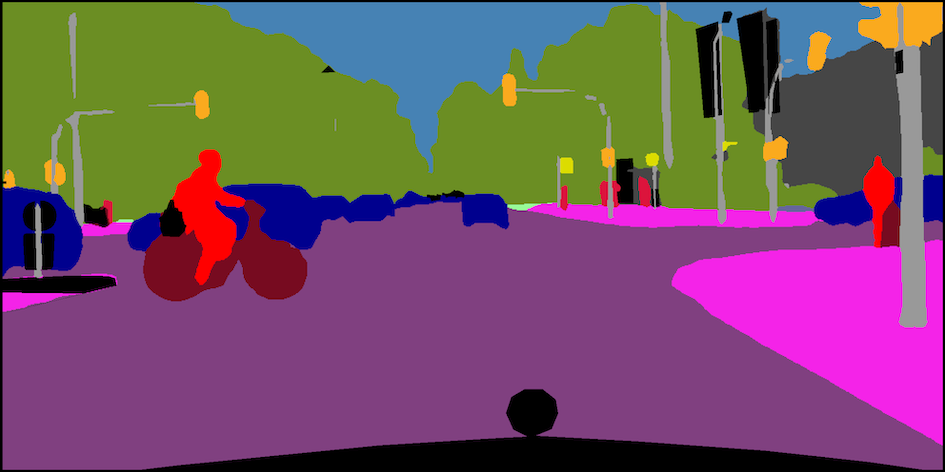} &
\includegraphics[width=0.24\linewidth]{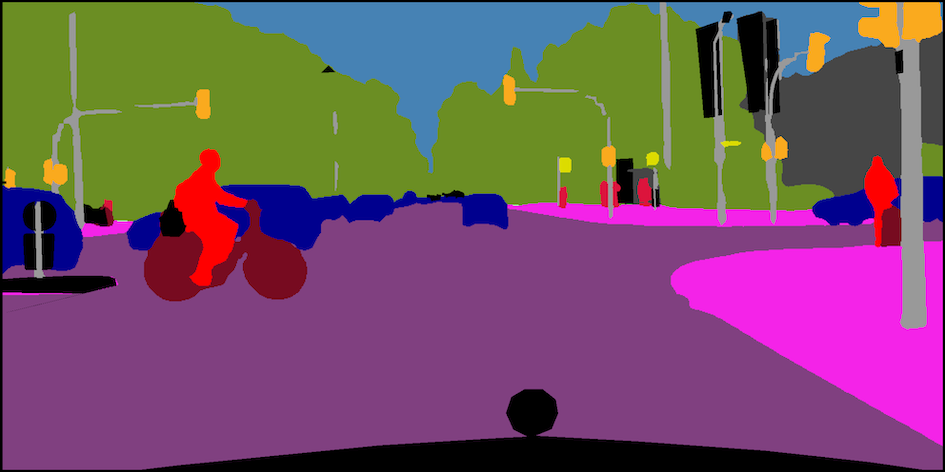} &
\includegraphics[width=0.24\linewidth]{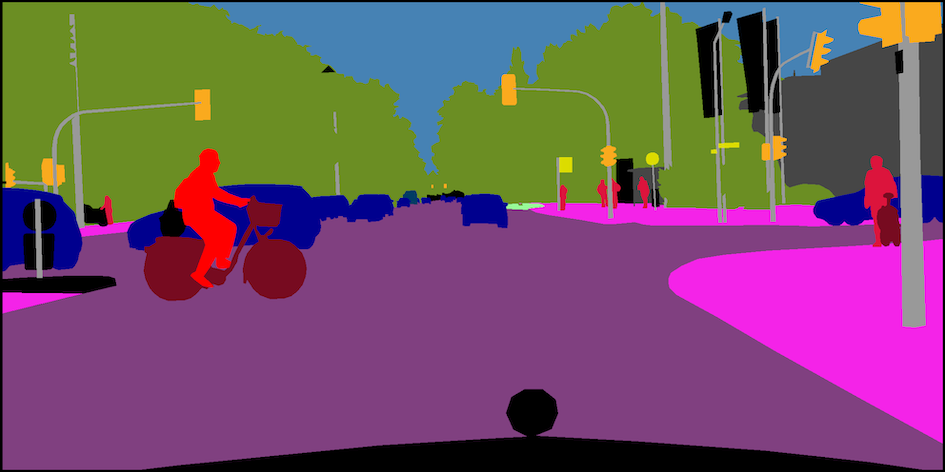} \\
\includegraphics[width=0.24\linewidth]{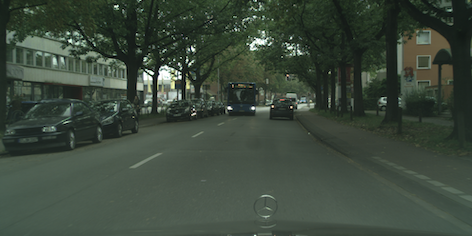} &
\includegraphics[width=0.24\linewidth]{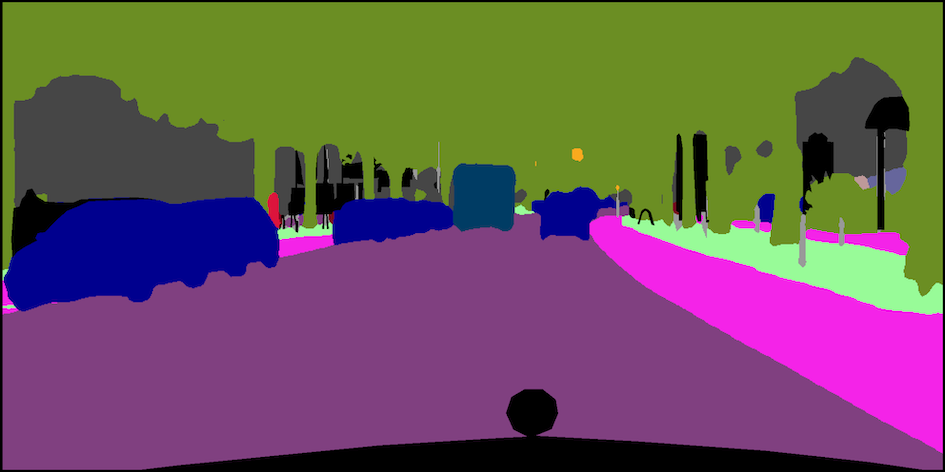} &
\includegraphics[width=0.24\linewidth]{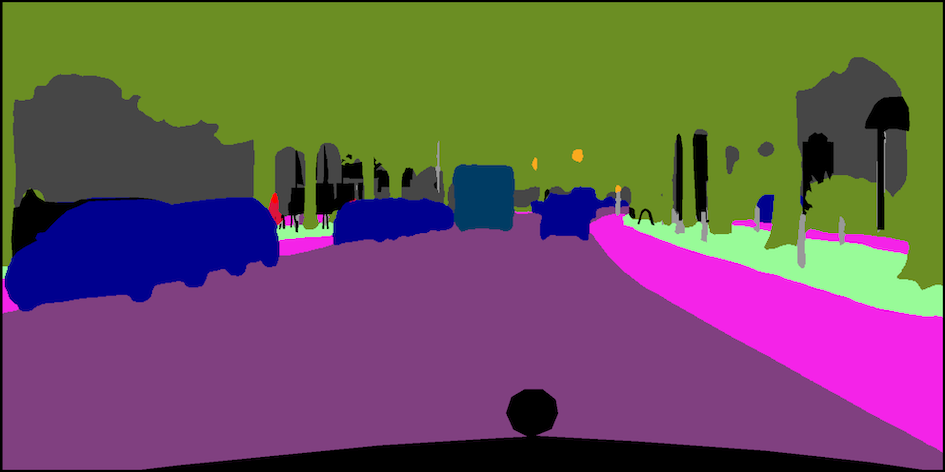} &
\includegraphics[width=0.24\linewidth]{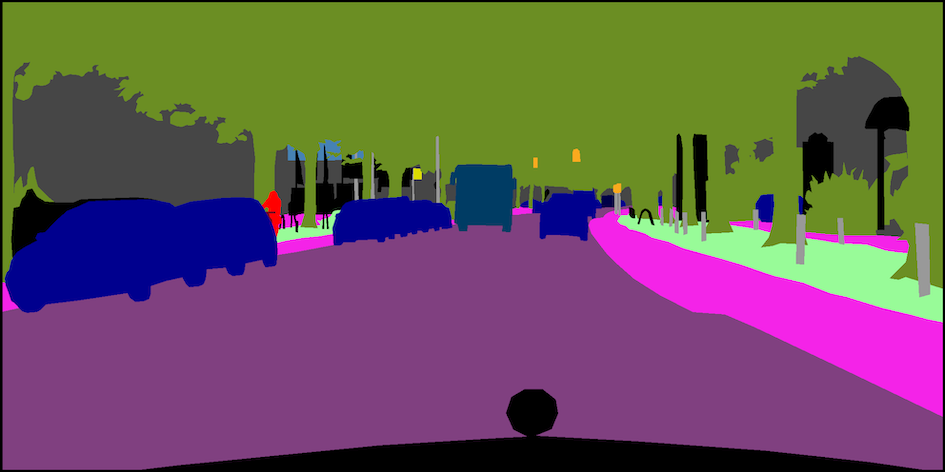} \\
(a) Input &
(b) BiSeNetV2 &
(c) BiSeNetV2-Large &
(d) Groundtruth \\
\end{tabular}
\end{center}
\caption{\textbf{Visualization examples on the Cityscapes \emph{val} set produced from BiSeNetV2 and BiSeNetV2-Large.} The first row shows that our architecture can focus on the details, \eg fence. The bus in the third row demonstrates the architecture can capture the large object. The bus in the last row illustrates the architecture can encode the spatial context to reason it. }
\label{fig:vis:cityscapes}
\end{figure*}

\paragraph{Cityscapes.}
We present the segmentation accuracy and inference speed of the proposed BiSeNetV2 on City\-scapes \emph{test} set.
We use the training set and validation set with $2048\times1024$ input to train our models, which is resized into $1024\times512$ resolution at first in the models.
Then we evaluate the segmentation accuracy on the test set.
The measurement of inference time is conducted on one NVIDIA GeForce 1080Ti card.
Table~\ref{tab:city-performance-comp} reports the comparison results of our method and state-of-the-art methods.
The first group is non-real-time methods, containing CRF-RNN~\citep{Zheng-ICCV-CRFasRNN-2015}, Deeplab-CRF \citep{Chen-ICLR-Deeplabv2-2016}, FCN-8S~\citep{Long-CVPR-FCN-2015}, Dilation10~\citep{Yu-ICLR-Dilate-2016}, LRR~\citep{Ghiasi-ECCV-LRR-2016}, Deeplabv2-CRF~\citep{Chen-Arxiv-Deeplabv2-2016}, FRRN~\citep{Pohlen-CVPR-FRRN-2017}, RefineNet~\citep{Lin-CVPR-Refinenet-2017}, DUC~\citep{Wang-WACV-DUC-2018}, PSPNet~\citep{Zhao-CVPR-PSPNet-2017}.
The real-time semantic segmentation algorithms are li\-sted in the second group, including ENet~\citep{Paszke-Arxiv-ENet-2016}, SQ~\citep{Treml-NIPSW-SQ-2016}, ESPNet~\citep{Mehta-ECCV-ESPNet-2018}, ESPNetV2~\citep{Mehta-CVPR-ESPNetV2-2019}, ERFNet~\citep{Romera-TITS-ERFNet-2018}, Fast-SCNN~\citep{Poudel-Arxiv-FastSCNN-2019}, ICNet \citep{Zhao-ECCV-ICNet-2018}, DABNet~\citep{Li-BMVC-DABNet-2019}, DFANet \citep{Li-CVPR-DFANet-2019}, GUN~\citep{Mazzini-BMVC-GUN-2018}, SwiftNet~\citep{Orsic-CVPR-SwiftNet-2019}, BiSeNetV1~\citep{Yu-ECCV-BiSeNet-2018}.
The third group is our methods with different levels of complexities.
As shown in Table~\ref{tab:city-performance-comp}, our method achieves $72.6\%$ mean IoU with $156$ FPS and yields $75.3\%$ mean IoU with $47.3$ FPS, which are \textit{state-of-the-art} trade-off between accuracy and speed.
These results are even better than several non-real-time algorithms in the second group of Table~\ref{tab:city-performance-comp}.
We note that many non-real-time methods may adopt some evaluation tricks, \eg multi-scale testing and multi-crop evaluation, which can improve the accuracy but are time-consuming.
Therefore, we do not adopt this strategy with the consideration of speed.
To better view, we illustrate the trade-off between performance and speed in Figure~\ref{fig:comparison}.
To highlight the effectiveness of our method, we also present some visual examples of BiSeNetV2 on Cityscapes in Figure~\ref{fig:vis:cityscapes}.

\paragraph{CamVid.}
Table~\ref{tab:camvid-performance-comp} shows the statistic accuracy and speed results on the CamVid dataset. 
In the inference phase, we use the training dataset and validation dataset to train our model with $960\times720$ resolution input. 
Our models are compared to some non-real-time algorithms, including SegNet~\citep{Badrinarayanan-PAMI-SegNet-2017}, Deeplab~\citep{Chen-ICLR-Deeplabv2-2016}, RTA~\citep{Huang-ECCV-RTA-2018}, Dilate8~\citep{Yu-ICLR-Dilate-2016}, PSPNet \citep{Zhao-CVPR-PSPNet-2017}, VideoGCRF~\citep{Chandra-CVPR-VideoGCRF-2018}, and DenseDecoder \citep{Bilinski-CVPR-DenseDecoder-2018}, and real-time algorithms, containing ENet~\citep{Paszke-Arxiv-ENet-2016}, ICNet~\citep{Zhao-ECCV-ICNet-2018}, DABNet~\citep{Li-BMVC-DABNet-2019}, DFANet~\citep{Li-CVPR-DFANet-2019}, SwiftNet~\citep{Orsic-CVPR-SwiftNet-2019}, BiSeNetV1~\citep{Yu-ECCV-BiSeNet-2018}.
BiSeNetV2 achieves much faster inference speed than other methods.
Apart from the efficiency, our accuracy results also outperform these work.
Besides, we investigate the effect of the pre-training datasets on CamVid.
The last two rows of Table~\ref{tab:camvid-performance-comp} show that pre-training on Cityscapes can greatly improve the mean IoU over $6\%$ on the CamVid test set.

\begin{table}[t]
\centering
\small
\tablestyle{2pt}{1.2}
\caption{\textbf{Comparison with state-of-the-art on COCO-Stuff}.
Our models are trained and evaluated with the input of $640\times640$ resolution.
Notation: \textit{backbone} is the backbone models pre-trained on ImageNet dataset.
}
\label{tab:stuff-performance-comp}
\begin{tabular}{l|c|c|c|c|c}
\shline
\multicolumn{1}{c|}{method} & ref. & backbone & mIoU(\%) & pixAcc(\%) & FPS \\ \hline
\multicolumn{5}{l}{\textit{large models}} \\ \hline
FCN-16s 	& CVPR2017 & VGG16     & 22.7  & 52.0 & 5.9 \\
Deeplab 	& ICLR2015 & VGG16     & 26.9  & 57.8 & 8.1 \\
FCN-8S      & CVPR2015 & VGG16 	& 27.2  & 60.4 & -   \\
PSPNet50    & CVPR2017 & ResNet50 	& \textbf{32.6}	& -    & 6.6 \\
 \hline
 \multicolumn{5}{l}{\textit{small models}} \\ \hline
ICNet 		& ECCV2018 & PSPNet50  & \underline{29.1}  & -    & 35.7 \\
\hline 
BiSeNetV2   & {---} & no  & 25.2 & \underline{60.5} &  \textbf{87.9} \\
BiSeNetV2-L & {---} & no  & 28.7 & \textbf{63.5}  & \underline{42.5} \\
\shline
\end{tabular}
\end{table}

\paragraph{COCO-Stuff.} 
We also report our accuracy and speed results on COCO-Stuff validation dataset in Table~\ref{tab:stuff-performance-comp}. 
In the inference phase, we pad the input into $640\times640$ resolution. 
For a fair comparison~\citep{Long-CVPR-FCN-2015, Chen-Arxiv-Deeplabv2-2016, Zhao-CVPR-PSPNet-2017, Zhao-ECCV-ICNet-2018}, we do not adopt any time-consuming testing tricks, such as multi-scale and flipping testing.
Even With more complex categories in this dataset, compared to pioneer work, our BiSeNetV2 still performs more efficient and achieve comparable accuracy.

\section{Concluding Remarks}
\label{sec:conclusion}
We observe that 
the semantic segmentation task requires both low-level details and high-level semantics.
We propose a new architecture to treat the spatial details and categorical semantics separately, termed Bilateral Segmentation Network (BiSeNetV2).
The BiSeNetV2 framework is a generic architecture, which can be implemented by 
most
convolutional models.
Our instantiations of BiSeNetV2 achieve a 
good 
trade-off between segmentation accuracy and inference speed.
We hope that this generic architecture BiSeNetV2 will foster further research in semantic segmentation.

\section*{Acknowledgment}
This work is supported by the National Natural Science Foundation of China (No. 61433007 and 61876210).

{\small
	\bibliographystyle{spbasic}
	\bibliography{reference}
}

\end{document}